\documentclass{article}
\usepackage{silence}
\usepackage[table]{xcolor}
\usepackage[a-2b]{pdfx}
\usepackage[preprint]{corl_2026}

\usepackage{pifont}
\usepackage{adjustbox}
\usepackage{capt-of}
\usepackage{makecell}
\usepackage{siunitx}
\usepackage{graphicx}
\usepackage{amsmath}
\usepackage{amssymb}
\usepackage{fvextra}
\usepackage[skins,breakable]{tcolorbox}
\input{figures/api/pygments.tex}
\fvset{breaklines=true, breakanywhere=false}
\definecolor{cellframe}{gray}{0.85}
\definecolor{cellback}{gray}{0.97}
\definecolor{cellcounter}{gray}{0.6}
\newtcolorbox{cellcode}[1]{enhanced, colback=cellback, colframe=cellframe, boxrule=0.4pt, arc=1.5pt, left=24pt, right=6pt, top=3pt, bottom=3pt, boxsep=0pt, before={\par\vspace{8pt}\noindent}, after skip=0pt, fontupper=\footnotesize, overlay={\node[anchor=north east, inner sep=0pt, font=\footnotesize\ttfamily, text=cellcounter] at ([xshift=19pt, yshift=-4.4pt]frame.north west) {[#1]};}}
\newtcolorbox{cellout}{blank, before skip=8pt, after skip=0pt, fontupper=\scriptsize}
\newenvironment{cellgroup}{\par\vspace{16pt}\noindent\begin{minipage}{\linewidth}}{\end{minipage}\par}

\usepackage{corl_2026}
\definecolor{checkgreen}{HTML}{06bf4d}
\definecolor{checkorange}{HTML}{bf8a06}
\definecolor{checkred}{HTML}{bf2206}
\newcommand{\customcheckmark}{\textcolor{checkgreen}{\textbf{\ding{51}}}}
\newcommand{\customorangecheckmark}{\textcolor{checkorange}{\textbf{\ding{51}}}}
\newcommand{\customnocheckmark}{\textcolor{checkred}{\textbf{\ding{55}}}}
\newcommand{\grayrow}{\rowcolor{gray!10}}

\title{\raisebox{-0.413em}{\includegraphics[height=1.328em]{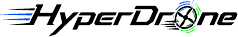}}: Simulation-to-Reality Transfer at Millions of Frames per Second}

\author{
  Jonas~Eschmann \quad Giuseppe Loianno
  \vspace{2pt}
  \\
  University of California, Berkeley
}

\begin{document}
\maketitle

\vspace{-7mm}
\begin{center}
\begin{tabular}{@{}r@{\textbf{:}\ }l@{}}
\textbf{Project Page} & \url{https://hyperdrone.dev} 
\vspace{0.8pt}
\\
\textbf{Video} & \url{https://youtu.be/rc3E-GuLqDw} 
\vspace{0.8pt}
\\
\textbf{Code}  & \url{https://github.com/rl-tools/hyperdrone}
\end{tabular}
\end{center}

\begin{quote}
    \centering
    \includegraphics[width=\linewidth]{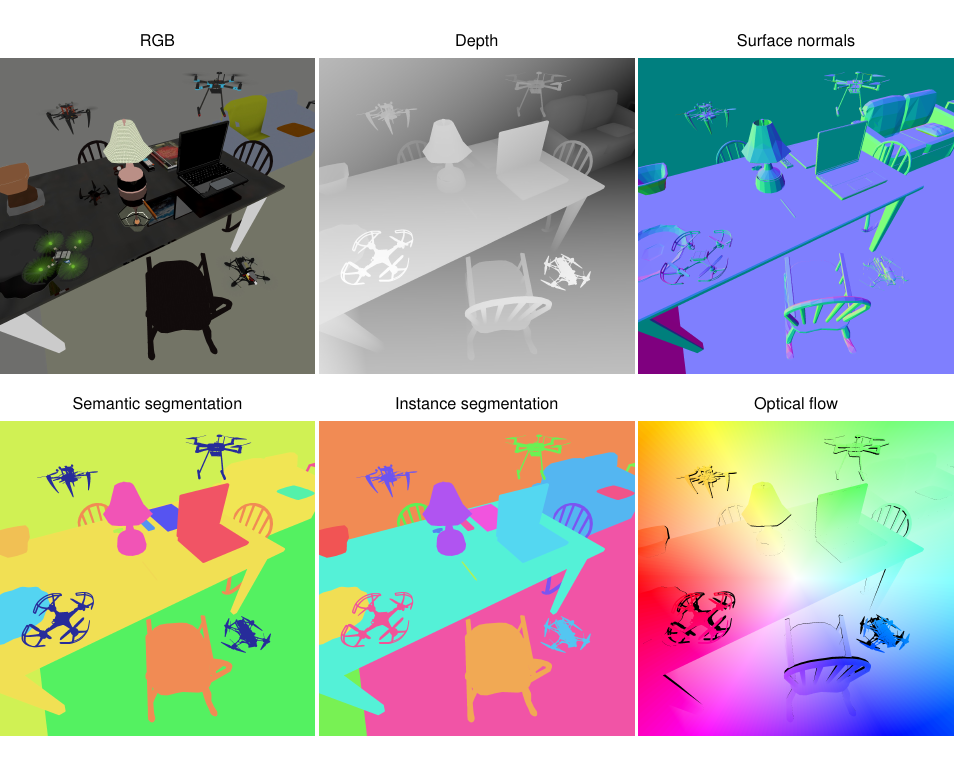}
    \vspace{-9mm}
    \captionof{figure}{HyperDrone Output Modalities.}
    \label{fig:modalities}
\end{quote}

\begin{abstract}
We present HyperDrone, an extremely fast quadrotor simulator focused on high-performance batch rendering that is more than $20\times$ faster than existing state-of-the-art batch quadrotor simulators. HyperDrone combines highly optimized multirotor dynamics with HyperLux, a new ray-tracing engine that supports Nvidia OptiX, Vulkan, Apple Metal and WebGPU backends that takes full advantage of modern ray-tracing hardware, capable of running both components entirely on the GPU. HyperLux renders RGB, depth, semantic/instance segmentation, surface normals and optical flow observations at configurable levels of fidelity, including reflections, realistic lighting, motion blur, anti-aliasing, and Physically Based Rendering (PBR). Even in the highest-fidelity settings, \mbox{HyperLux} achieves millions of Frames Per Second (FPS) on consumer-grade GPUs while rendering realistic 3D scenes. We demonstrate the merits of HyperDrone by training a vision-based yaw-control policy purely in simulation and transferring it to a real small-scale quadrotor (Crazyflie) with a camera, running completely onboard. 
\end{abstract}


\section{Introduction}
In recent years, imitation learning from human demonstrations has become a dominant method for advanced learning-based control of robots. This is particularly the case in the field of manipulation \cite{chi2023diffusionpolicy, chi2024diffusionpolicy,Ze2024DP3} and locomotion \cite{2018-TOG-deepMimic, manilocsurvey}. Conversely, for aerial vehicle autonomy, imitation learning from human demonstrations has not experienced comparable widespread adoption. In the case of quadrotors, in particular, two major trends have been imitation learning from (privileged) expert policies \cite{loquercio_2021_learning_high_speed_flight_in_the_wild,kaufmann2020RSS,eschmann_2026_raptor} and using pre-trained models as vision encoders \cite{saviolo2025nova,xing2024bootstrapping}. The latter promise to bring broader generalization by virtue of having been pre-trained on massive amounts of real-world images collected from the internet.

Imitation learning from human demonstrations and pre-trained vision models are motivated and driven by scaling laws that were first observed in Large Language Models (LLMs) \cite{kaplan2020scaling}. Early indications hint that similar scaling laws might exist in robot imitation learning \cite{lin2024data}. The potential that scaling robot imitation data could bear similarly large returns as LLM scaling, has led to the formation of a large number of commercial robot imitation data creation/curation companies. These companies also bet on monopolizing forces known as the ``data flywheels'' \cite{grannen2025robot} limiting accessibility while the capital expenditure (CAPEX) requirement for large-scale robot data collection is by itself already a major hurdle for reproducibility and progress in the community.

The alternative is training robot control policies entirely in simulation without requiring real-world data. Not only are methods fully trained on synthetic data more accessible to fellow researchers, but they are also more introspectable. Due to the lack of interpretability of deep neural networks, in general we cannot get strong guarantees for learned end-to-end controllers. The best way to reason about the behavior of the trained control policy is through introspection of the data that went into training it. For vision foundation models \cite{dosovitskiy2021an, radford2021learning, simeoni2025dinov3, carion2025sam3segmentconcepts} this data is usually proprietary, hampering reproducibility, interpretability of the resulting behavior and the opportunity for independent researchers to study effects of modifying the data distribution.

To accelerate the research of vision-based aerial vehicle control we present the following contributions:
\vspace{-2mm}
\begin{enumerate}
\item \textbf{HyperDrone}: An extremely fast simulator for training end-to-end visual control policies. HyperDrone consists of:
\begin{itemize}
\item \textbf{HyperLux}: A fast, large-batch ray-tracing renderer for RGB, depth, semantic/instance segmentation, surface normals and optical flow perception
\item Highly optimized multirotor dynamics simulation
\end{itemize}
\item \textbf{Sim2Real Transfer} of a yaw-control policy trained entirely in simulation
\end{enumerate}
\vspace{-2mm}
\section{Related Work}
\label{sec:related-work}

\begin{table*}[b]
\vspace{-10pt}
\centering
\def\arraystretch{1.5}
\begin{adjustbox}{width=1.0\linewidth}
\setlength{\tabcolsep}{1mm}
\begin{tabular}{c c | c c c c c c c c c }
\hline
\textbf{Name} & \textbf{Engine} &
\textbf{\rule[-2.5ex]{0pt}{6.5ex} Quadrotor} &
\makecell{\textbf{Batch}\\\textbf{Rendering}} &
\makecell{\textbf{Blackwell}\\\textbf{Compatible}} &
\makecell{\textbf{Ray-tracing}\\\textbf{(Cores)}} &
\makecell{\textbf{Motion}\\\textbf{Blur}} &
\makecell{\textbf{Anti}\\\textbf{Aliasing}} &
\textbf{Reflections} &
\makecell{\textbf{Maintained}\\\textbf{Dependencies}} &
\textbf{Linux/Windows} \\
\hline

\grayrow
Flightmare \cite{song2020flightmare} & Unity &
\customcheckmark & \customnocheckmark & \customcheckmark & \customnocheckmark \; (\customnocheckmark) & \customnocheckmark & \customnocheckmark & \customnocheckmark & \customnocheckmark & \customcheckmark / \customnocheckmark \\

VisFly \cite{li2025visfly} & Habitat &
\customcheckmark & \customcheckmark & \customcheckmark & \customnocheckmark  \; (\customnocheckmark) & \customnocheckmark & \customnocheckmark & \customnocheckmark & \customcheckmark & \customcheckmark / \customnocheckmark \\

\grayrow
Aerial Gym \cite{kulkarni2025aerial} & Isaac Gym &
\customcheckmark & \customcheckmark & \customnocheckmark & \customcheckmark  \; (\customnocheckmark) & \customnocheckmark & \customnocheckmark & \customnocheckmark & \customnocheckmark & \customcheckmark / \customnocheckmark \\

OmniDrones \cite{xu2024omnidrones} & Isaac Lab &
\customcheckmark & \customcheckmark & \customnocheckmark & \customnocheckmark  \; (\customnocheckmark) & \customnocheckmark & \customcheckmark & \customnocheckmark & \customnocheckmark & \customcheckmark / \customnocheckmark \\

\grayrow
Madrona Ray-tracing \cite{rosenzweig2024madrona} & Madrona &
\customnocheckmark & \customcheckmark & \customcheckmark & \customcheckmark  \; (\customnocheckmark) & \customnocheckmark & \customnocheckmark & \customnocheckmark & \customcheckmark & \customcheckmark / \customcheckmark \\

Madrona Rasterization \cite{rosenzweig2024madrona} & Madrona &
\customnocheckmark & \customcheckmark & \customcheckmark & \customnocheckmark  \; (\customnocheckmark) & \customnocheckmark & \customnocheckmark & \customnocheckmark & \customcheckmark & \customcheckmark / \customcheckmark \\

\grayrow
\textbf{HyperDrone} Low & HyperDrone &
\customcheckmark & \customcheckmark & \customcheckmark & \customcheckmark  \; (\customcheckmark) & \customcheckmark & \customcheckmark & \customnocheckmark & \customcheckmark & \customcheckmark / \customcheckmark \\

\textbf{HyperDrone} Medium & HyperDrone &
\customcheckmark & \customcheckmark & \customcheckmark & \customcheckmark  \; (\customcheckmark) & \customcheckmark & \customcheckmark & \customorangecheckmark & \customcheckmark & \customcheckmark / \customcheckmark \\

\grayrow
\textbf{HyperDrone} High & HyperDrone &
\customcheckmark & \customcheckmark & \customcheckmark & \customcheckmark  \; (\customcheckmark) & \customcheckmark & \customcheckmark & \customcheckmark & \customcheckmark & \customcheckmark / \customcheckmark \\
\end{tabular}
\end{adjustbox}
\caption{Comparison of quadrotor simulators and rendering backends.}
\label{tab:simulator_comparison}
\end{table*}

In the recent past, several quadrotor simulators have been proposed and optimized for different purposes. We focus on simulating cameras and quadrotor dynamics to support visual control policies, and the main related simulators are listed in Table \ref{tab:simulator_comparison}. 

A very popular simulator for this purpose has been Flightmare \cite{song2020flightmare}. However, it does not support batch rendering and hence is comparatively slow. It has been measured to yield $\approx 230$ Frames Per Second (FPS) in \cite{li2025visfly} which is prohibitively slow for reinforcement learning which usually requires millions or even billions of samples. Furthermore, it requires proprietary, licensed software to replace the default scenes. 

To overcome the low rendering throughput, multiple batch-rendering-based simulators have been proposed. These cluster into Nvidia Isaac-based and Habitat-based simulators. VisFly \cite{li2025visfly} is based on a custom version of Habitat \cite{habitat19iccv} for batch rendering. Habitat in turn uses OpenGL through the Magnum engine. It is the only other simulator that supports quadrotor dynamics, batch rendering and is compatible with the latest generation GPUs (Nvidia Blackwell/5000-series, released in early 2025). However, it falls short of HyperDrone when it comes to fidelity features like motion blur and anti-aliasing. Note that we leave an exhaustive performance comparison for later and only focus on qualitative features in this section.

Aerial Gym \cite{kulkarni2025aerial} is based on Isaac Gym and implements a batch raytracer mainly for depth perception and semantic segmentation. While depth perception and semantic segmentation is also supported in HyperDrone, our main focus is on RGB observations which are not supported for general 3D mesh scenes (only for simple geometric objects) by Aerial Gym. Furthermore, due to Nvidia Warp limitations, Aerial Gym does not take advantage of ray-tracing cores in modern GPUs. Additionally, Isaac Gym, which Aerial Gym builds upon, was deprecated in mid-2024 and hence Aerial Gym does not support Nvidia Blackwell GPUs (5000 series). 

OmniDrones \cite{xu2024omnidrones} follows a similar strategy by building onto Isaac Sim. Compared to Aerial Gym it does not implement its own rendering pipeline and relies on Isaac Sim for camera simulation. OmniDrones is built on Isaac Sim 4.1.0 which was released in mid-2024 and hence does not support Blackwell/5000-series GPUs, either. Since the project is not maintained anymore we do not expect this to change. 

Initially, when conceptualizing HyperDrone we were planning to build it on top of Madrona \cite{rosenzweig2024madrona}, an open-source batch raytracer and rasterizer that has shown promising performance. However, while investigating Madrona we found that it does not take advantage of ray-tracing hardware in RTX-series cards and that the codebase is relatively complex. 

Hence, we concluded that there is a need for a new simulator that focuses on rendering performance while supporting the needs of realistic visual quadrotor simulation like motion blur, anti-aliasing and reflections. While designing HyperDrone we also keep in mind that in the future we should support rolling-shutter simulation as well, even though for now we focus on global shutter cameras.

\section{Method}
In the following we describe the architecture of HyperDrone and then the control problem formulation, platform and general setup for the simulation-to-reality transfer of a visual policy trained from scratch in HyperDrone.
\subsection{Architecture}

\begin{figure}
    \centering
    \includegraphics[width=0.70\textwidth]{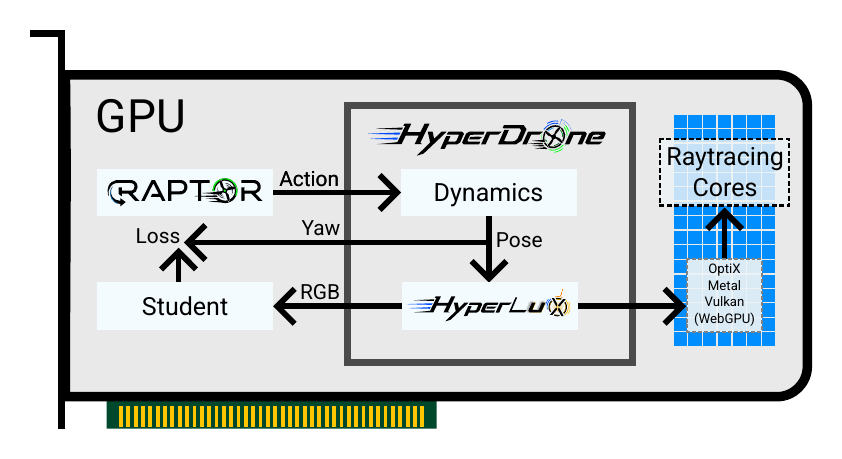}
    \vspace{-5mm}
    \caption{HyperDrone Architecture.}
    \label{fig:gpu}
    \vspace{-5mm}
\end{figure}
A main premise guiding the architecture of HyperDrone is that we want to avoid CPU $\leftrightarrow$ GPU memory transfers that commonly occur when quadrotor dynamics simulation is being run on the CPU while a batch renderer is run on the GPU. With OptiX, both run on the GPU. Another premise motivated in Section \ref{sec:related-work} is the ability to take advantage of the ray-tracing cores in modern hardware. 

The first component of HyperDrone is HyperLux, our ray-tracing renderer. Ray-tracing cores are fixed-function hardware that accelerate finding intersections of 3D rays with a 3D mesh. Compared to rasterization, which maps 3D geometry from a 3D mesh onto the camera plane, ray-tracing takes the more physically motivated approach of tracing rays of light. In practice, we shoot a ray from each pixel towards the 3D mesh and find the first point of intersection (this is what the ray-tracing fixed-function hardware supports). Once we identify the intersection, we can use the information from the mesh to e.g. look up the appropriate color for that particular intersection point from a texture map.
To make this tractable the 3D mesh first needs to be processed into an accelerated data-structure. This acceleration structure and using ray-tracing cores to intersect rays with it is not accessible via compute APIs like CUDA. HyperLux therefore uses Nvidia OptiX, Vulkan or Apple Metal, which handle the acceleration structure and return intersection points for the rays we supply, and falls back to software ray traversal on WebGPU, which does not expose ray-tracing hardware. 

The second component of HyperDrone is the quadrotor dynamics simulator. The quadrotor dynamics simulation is based on the implementation introduced in \cite{eschmann2024learningtofly} combined with years of continuous development, refactoring and improvements.
Due to rendering not being the bottleneck anymore (cf. rendering performance benchmarks in Section \ref{sec:results}), we can just render every frame from scratch. This basically rules out overfitting to any particular dataset in the traditional sense. We might still overfit to the data-distribution that the simulator generates (e.g. in terms of visual fidelity, artifacts etc.) but overfitting to particular samples is virtually ruled out by never using a sample twice.

For the proof-of-concept of the simulation-to-reality transfer, we use RAPTOR \cite{eschmann_2026_raptor} as a teacher policy and train a student policy to estimate the yaw between a target frame and the current frame (more in the next Section \ref{sec:sim2real}). We run both the teacher and the student on the GPU as well. Please refer to Figure~\ref{fig:gpu} for a graphical overview of the HyperDrone architecture.

\subsection{Visual Simulation-to-Reality Transfer}
\label{sec:sim2real}
\begin{figure}
    \centering
    \includegraphics[width=0.80\textwidth]{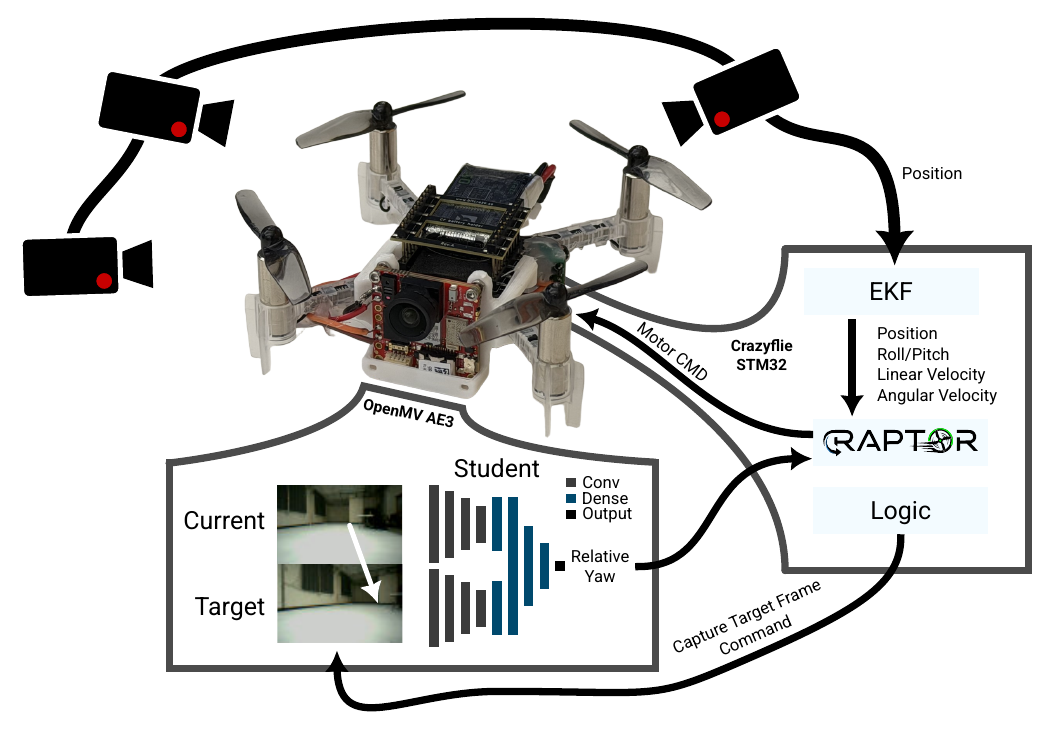}
    \vspace{-3mm}
    \caption{Control Architecture Overview.}
    \label{fig:experimental-setup}
    \vspace{-3mm}
\end{figure}
\begin{figure}[b]
    \vspace{-7mm}
    \centering
    \begin{minipage}[t]{0.48\linewidth}
        \centering
        \includegraphics[width=0.6\linewidth]{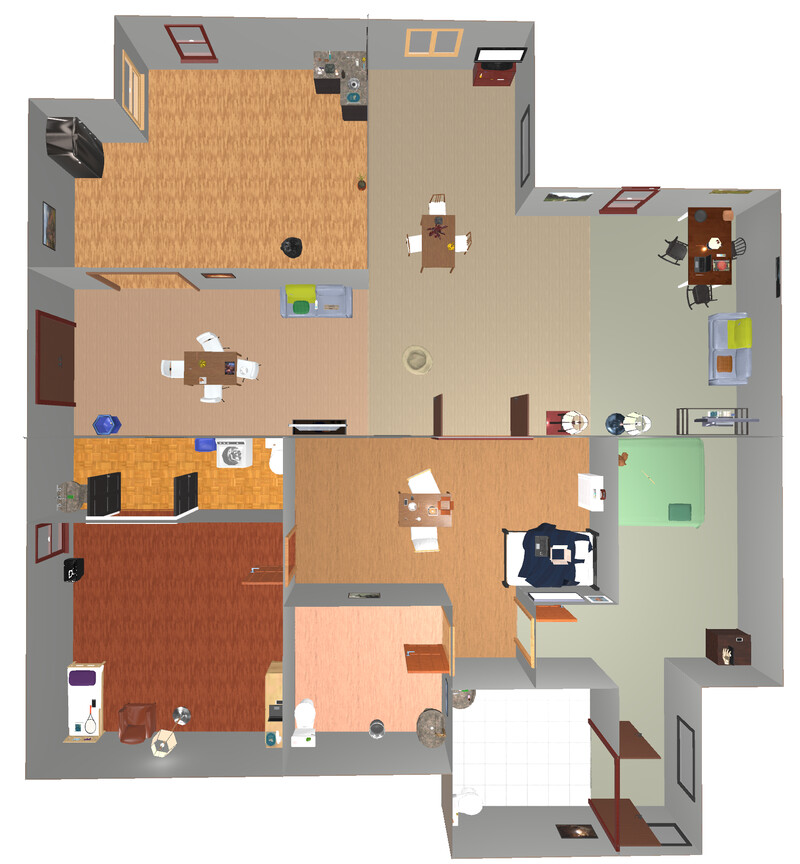}
        \vspace{-1mm}
        \caption{ProcTHOR \cite{procthor} Train-1 Scene.}
        \label{fig:procthor}
    \end{minipage}
    \hfill
    \begin{minipage}[t]{0.48\linewidth}
        \centering
        \includegraphics[width=\linewidth]{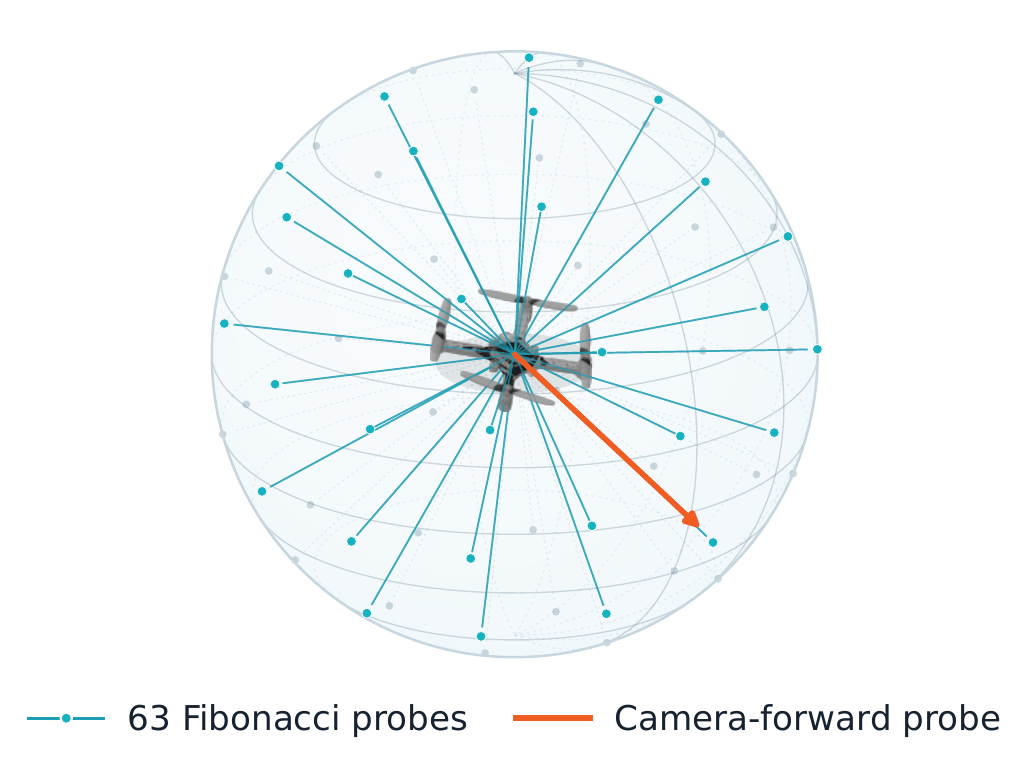}
        \vspace{-5.2mm}
        \caption{Probe Rays.}
        \label{fig:probe-rays}
    \end{minipage}
\end{figure}

To show that HyperDrone actually allows simulation-to-reality transfer after training from scratch on synthetic images, we propose a simple task, a training pipeline and a hardware platform to deploy it on and an experimental setup to test the trained network. 

We choose a simple task to disentangle the visual simulation-to-reality transfer from control challenges. As can be seen in Figure \ref{fig:experimental-setup}, we abstract most of the high-frequency control problem by using the RAPTOR \cite{eschmann_2026_raptor} foundation policy for low-level and position control and only leave yaw control to the network trained in simulation and deployed on a real camera. The task of the student policy is to provide the relative yaw offset between a target frame and the current frame. Hence, we use a convolutional network on the channel-wise stack of both frames, followed by a dense head that predicts the relative yaw.

\begin{figure}
    \centering
    \includegraphics[width=1.0\textwidth]{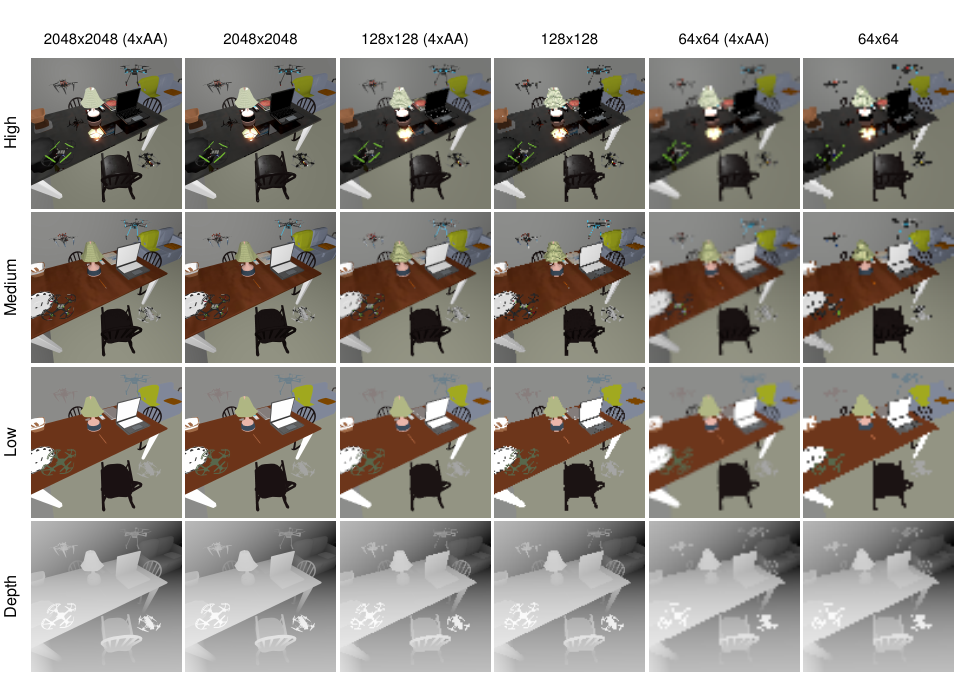}
    \vspace{-9mm}
    \caption{Rendering Fidelity Comparison with Four Secondary Drones.}
    \label{fig:fidelity}
    \vspace{-6mm}
\end{figure}

\begin{figure}
    \centering
    \includegraphics[width=0.95\textwidth]{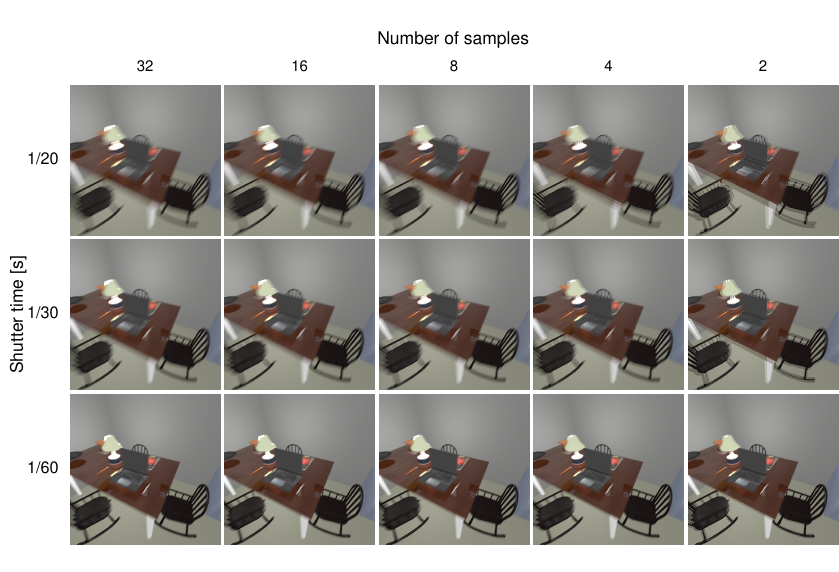}
    \vspace{-5mm}
    \caption{Motion Blur Parameters.}
    \label{fig:motion-blur}
    \vspace{-6mm}
\end{figure}

We want to support a large diversity of 3D scenes to avoid overfitting to a specific scene. Hence, we use the procedurally generated indoor scenes from the ProcTHOR \cite{procthor} dataset. This dataset includes $10000$ diverse indoor scenes with randomized room layouts and object placements (cf. Figure \ref{fig:procthor}). It also strikes a good balance between realism and performance (cf. Figures \ref{fig:fidelity} and \ref{fig:motion-blur}). However, HyperDrone is not limited to ProcTHOR, it can load general binary glTF 2.0 (\texttt{.glb}) files. And glTF is a widely used open standard, also known as the ``JPEG of 3D'' \cite{hevele2022glb}. Supporting open and widely used formats is a large advantage for the community accessibility of HyperDrone over other simulators like Flightmare which rely on proprietary formats. 

Given a 3D scene mesh, we need to find random positions that can be used as initial states for our training pipeline. Since the room layouts are irregularly shaped (cf. Figure \ref{fig:procthor}) and also because there are random obstacles and walls in the scenes, we cannot use simple uniform position sampling. Hence, we implement a probe-ray-based rejection sampling mechanism that allows us to specify the minimum distance to any object. As depicted in Figure \ref{fig:probe-rays}, we use 64 probe rays that are shot out from the ego-camera position in all directions using a deterministic Fibonacci sphere distribution. These probe rays use the same ray-tracing hardware acceleration as the rays used to retrieve pixel colors and hence the performance overhead is minimal (roughly $\frac{64}{64\times 64} \approx 1.6\%$ for $64\times64$ frames, even lower for higher resolutions). In the future, these probe rays can also be used for collision detection with the actual 3D mesh, not some simplified collision mesh as in most other simulators/engines.

For control, RAPTOR \cite{eschmann_2026_raptor} outputs direct motor commands at $500$ Hz, based on position observations from a motion capture system as well as roll/pitch, linear velocity and angular velocity observations from the on-board Extended Kalman Filter (EKF) on the Crazyflie platform. 
We mount an OpenMV AE3 camera board on the Crazyflie and run the trained neural network on a Neural Processing Unit (NPU) on the camera board. We develop a Universal Asynchronous Receiver/Transmitter (UART)-based protocol to transfer the yaw commands to the Crazyflie microcontroller to be consumed by RAPTOR. 

The protocol also allows for a backchannel that allows instructing the inference code to take the current image as a target image. The ``Logic'' corresponds to a simple state machine that first instructs the capturing of two target frames with a yaw offset between them. After that, the packet contains another bit that can select between two target frames which leads to the back-and-forth yaw control that you can see in the complementary video and in Figure \ref{fig:yaw-tracking}. This demonstrates the successful simulation-to-real transfer of a visual control policy that has been only trained on synthetic images from HyperDrone.

\section{Results}
\label{sec:results}

In this section, we first investigate the visual results, analyzing the effects of different parameters in the HyperDrone raytracer. Then we analyze the performance, and finally we discuss results from the simulation-to-reality transfer.

\subsection{Visual Results}
In Figure \ref{fig:fidelity}, we compare rendering modes, resolutions and anti-aliasing. The comparison includes three ray-tracing modes (more details in Appendix~\ref{appendix:rendering_fidelity_modes}):
\vspace{-2mm}
\begin{enumerate}
\item \textbf{High}: Supports complex reflections like the reflection of the picture on the wall in the table surface in Figure \ref{fig:fidelity}. It also supports more complex lighting (cf. desk lamp in Figure \ref{fig:fidelity}) and higher fidelity material appearance.
\item \textbf{Medium}: Still supports complex materials but has more limited support for reflections and lighting.
\item \textbf{Low}: No textures, pre-baked colors, optimized for performance.
\end{enumerate}
\vspace{-2mm}
Additionally, due to the popularity in the robot learning community \cite{loquercio_2021_learning_high_speed_flight_in_the_wild, kulkarni2024reinforcement,vijay2025vit, zhang2026highspeedvisionbasedflightclutter}, we also support depth, semantic/instance segmentation, surface normals and optical flow perception. Figure \ref{fig:modalities} shows these outputs alongside RGB rendered at high fidelity (cf. Appendix~\ref{appendix:rendering_fidelity_modes}). In Figure \ref{fig:fidelity}, we show different resolutions from $2048\times2048$ to $64\times64$ with anti-aliasing on and off. Particularly when comparing the lower resolution cases like in $64\times64$, we can see that there are strong aliasing effects. These effects happen because our ray is infinitesimally narrow and either hits an object or not, hence e.g. the edge of the table being hit in some pixels while others hit the ground. Naturally, this does not occur because each pixel collects photons from a frustum (a rectangular cone), densely sensing all the surfaces in the frustum and then averaging them. We implement anti-aliasing by simulating this natural process with $2\times2$ rays per pixel that are averaged, leading to vastly reduced aliasing. While more apparent for us at lower resolutions, even at higher resolutions aliasing is an unnatural effect that the neural networks could overfit to.

For fast-moving robots, particularly robots with agile rotational dynamics like quadrotors, it is important to take into account motion blur. Motion blur appears because the assumption that the camera takes a snapshot at a particular instant in time is not true. Cameras need to accumulate photons over time to build an accurate picture. While the shutter is open and the camera sensor is accumulating photons, the robot can move, leading to different objects contributing to any particular pixel over time. In Figure \ref{fig:motion-blur}, we show motion blur for different shutter durations. In contrast to game engines, where approximations are more common, we implement motion blur by simulating the actual physical process i.e. by subsampling $N$ rays throughout the shutter period and then averaging them for the final frame. In Figure \ref{fig:motion-blur}, the reader can see that just using $N=2$ subsamples leads to a relatively unrealistically appearing motion blur, particularly in the $1/20$ s shutter time case. When increasing the number of samples, the fidelity of the motion-blur simulation improves with $N=32$ showing realistic motion blur. Practically, we find that $N=2$ is sufficient for simulation-to-reality transfer (cf. Section \ref{sec:results-sim2real}).

\begin{table*}
\def\arraystretch{1.2}
\centering
\resizebox{\linewidth}{!}{%
\begin{tabular}{l | c | c c | c c | c c | c c}
\multicolumn{2}{c|}{} & \multicolumn{4}{c|}{20 Objects} & \multicolumn{4}{c}{ProcTHOR} \\
\cline{3-10}
\multicolumn{2}{l|}{} & \multicolumn{2}{c|}{RGB} & \multicolumn{2}{c|}{Depth} & \multicolumn{2}{c|}{RGB} & \multicolumn{2}{c}{Depth} \\
\hline
Task & \makecell{\#envs} & \makecell{\rule[-2.5ex]{0pt}{6.5ex} Render} & \makecell{Physics \&\\Render} & \makecell{Render} & \makecell{Physics \&\\Render} & \makecell{Render} & \makecell{Physics \&\\Render} & \makecell{Render} & \makecell{Physics \&\\Render} \\
\hline
\multicolumn{5}{l}{Nvidia RTX 4090 16GB (Mobile, 80W, OptiX)} & \multicolumn{5}{c}{} \\
\hline
\grayrow
  Aerial Gym & \num{256} & \num{4771} & \num{4347} & \num{126024} & \num{68651} & -- & -- & \num{133.32} & \num{128.32} \\
  Aerial Gym & \num{4096} & \num{1375} & \num{1328} & \num{127397} & \num{107428} & -- & -- & \num{130.97} & OOM \\
\grayrow
  OmniDrones & \num{8} & \num{359.33} & \num{295.97} & \num{371.90} & \num{299.13} & \num{276.53} & \num{233.78} & \num{297.27} & \num{246.07} \\
  OmniDrones & \num{4096} & T/O & T/O & T/O & T/O & OOM & OOM & OOM & OOM \\
\grayrow
  VisFly & \num{64} & \num{16731} & \num{3901} & \num{17773} & \num{3986} & \num{2818} & \num{1851} & \num{4871} & \num{2637} \\
  VisFly & \num{4096} & \num{15192} & \num{5915} & \num{15677} & \num{5929} & \num{2780} & \num{2221} & \num{4723} & \num{3302} \\
\grayrow
  Madrona Raytracing & \num{4096} & \num{366611} & -- & \num{366472} & -- & \num{283255} & -- & \num{283463} & -- \\
  Madrona Rasterization & \num{4096} & \num{1300710} & -- & \num{1295400} & -- & \num{94142} & -- & \num{94112} & -- \\
\grayrow
  \textbf{HyperDrone} Low & \num{4096} & \textbf{\num{2582870}} & \textbf{\num{2547080}} & \textbf{\num{2745197}} & \textbf{\num{2718340}} & \textbf{\num{1749660}} & \textbf{\num{1735410}} & \textbf{\num{1685167}} & \textbf{\num{1675053}} \\
  \textbf{HyperDrone} Medium & \num{4096} & \num{2193120} & \num{2158480} & \textbf{\num{2745197}} & \textbf{\num{2718340}} & \num{1296250} & \num{1285360} & \textbf{\num{1685167}} & \textbf{\num{1675053}} \\
\grayrow
  \textbf{HyperDrone} High & \num{4096} & \num{2082940} & \num{2068700} & \textbf{\num{2745197}} & \textbf{\num{2718340}} & \num{569534} & \num{564210} & \textbf{\num{1685167}} & \textbf{\num{1675053}} \\
\hline
\multicolumn{5}{l}{Apple iPhone 17 Pro Max 12GB (Mobile, $\sim$10W, Metal)} & \multicolumn{5}{c}{} \\
\hline
\grayrow
  \textbf{HyperDrone} Low & \num{4096} & \textbf{\num{531124}} & \textbf{\num{477410}} & \textbf{\num{506573}} & \textbf{\num{468499}} & \textbf{\num{174052}} & \textbf{\num{166105}} & \textbf{\num{175860}} & \textbf{\num{167240}} \\
  \textbf{HyperDrone} Medium & \num{4096} & \num{459535} & \num{424093} & \textbf{\num{506573}} & \textbf{\num{468499}} & \num{140696} & \num{134536} & \textbf{\num{175860}} & \textbf{\num{167240}} \\
\grayrow
  \textbf{HyperDrone} High & \num{4096} & \num{433191} & \num{403635} & \textbf{\num{506573}} & \textbf{\num{468499}} & \num{97376} & \num{93747} & \textbf{\num{175860}} & \textbf{\num{167240}} \\
\hline
\multicolumn{5}{l}{Apple MacBook Pro M3 Pro 36GB (Mobile, $\sim$45W, Metal)} & \multicolumn{5}{c}{} \\
\hline
\grayrow
  \textbf{HyperDrone} Low & \num{4096} & \textbf{\num{507150}} & \textbf{\num{503409}} & \textbf{\num{515534}} & \textbf{\num{512177}} & \textbf{\num{303754}} & \textbf{\num{302352}} & \textbf{\num{293296}} & \textbf{\num{288189}} \\
  \textbf{HyperDrone} Medium & \num{4096} & \num{473069} & \num{469994} & \textbf{\num{515534}} & \textbf{\num{512177}} & \num{223876} & \num{222955} & \textbf{\num{293296}} & \textbf{\num{288189}} \\
\grayrow
  \textbf{HyperDrone} High & \num{4096} & \num{480795} & \num{470713} & \textbf{\num{515534}} & \textbf{\num{512177}} & \num{147466} & \num{148880} & \textbf{\num{293296}} & \textbf{\num{288189}} \\
\hline
\multicolumn{5}{l}{Nvidia RTX 5060 Ti 16GB (Desktop, 180W, OptiX)} & \multicolumn{5}{c}{} \\
\hline
\grayrow
  \textbf{HyperDrone} Low & \num{4096} & \textbf{\num{2127682}} & \textbf{\num{2110431}} & \textbf{\num{2048953}} & \textbf{\num{2031962}} & \textbf{\num{1595207}} & \textbf{\num{1586442}} & \textbf{\num{1562579}} & \textbf{\num{1555133}} \\
  \textbf{HyperDrone} Medium & \num{4096} & \num{1790695} & \num{1779399} & \textbf{\num{2048953}} & \textbf{\num{2031962}} & \num{1111825} & \num{1106211} & \textbf{\num{1562579}} & \textbf{\num{1555133}} \\
\grayrow
  \textbf{HyperDrone} High & \num{4096} & \num{1785236} & \num{1773621} & \textbf{\num{2048953}} & \textbf{\num{2031962}} & \num{564196} & \num{563648} & \textbf{\num{1562579}} & \textbf{\num{1555133}} \\
\hline
\multicolumn{2}{l}{Nvidia RTX 5090 32GB (Desktop, 575W, OptiX)} & \multicolumn{8}{c}{} \\
\hline
\grayrow
  \textbf{HyperDrone} Low & \num{4096} & \textbf{\num{10723500}} & \textbf{\num{10354100}} & \textbf{\num{11204300}} & \textbf{\num{10919367}} & \textbf{\num{7574240}} & \textbf{\num{7449630}} & \textbf{\num{7739323}} & \textbf{\num{7530700}} \\
  \textbf{HyperDrone} Medium & \num{4096} & \num{9009180} & \num{8732240} & \textbf{\num{11204300}} & \textbf{\num{10919367}} & \num{5353660} & \num{5245080} & \textbf{\num{7739323}} & \textbf{\num{7530700}} \\
\grayrow
  \textbf{HyperDrone} High & \num{4096} & \num{8484850} & \num{8229860} & \textbf{\num{11204300}} & \textbf{\num{10919367}} & \num{2444850} & \num{2428840} & \textbf{\num{7739323}} & \textbf{\num{7530700}} \\
\hline
\multicolumn{5}{l}{Nvidia RTX 5090 32GB (Desktop, 575W, Vulkan)} & \multicolumn{5}{c}{} \\
\hline
\grayrow
  \textbf{HyperDrone} Low & \num{4096} & \num{11390406} & \textbf{\num{5540902}} & \textbf{\num{11543865}} & \textbf{\num{5531951}} & \num{6555618} & \textbf{\num{5331847}} & \textbf{\num{6626167}} & \textbf{\num{5332955}} \\
  \textbf{HyperDrone} Medium & \num{4096} & \num{11453317} & \num{5534717} & \textbf{\num{11543865}} & \textbf{\num{5531951}} & \textbf{\num{6790344}} & \num{5302185} & \textbf{\num{6626167}} & \textbf{\num{5332955}} \\
\grayrow
  \textbf{HyperDrone} High & \num{4096} & \textbf{\num{11489513}} & \num{5524774} & \textbf{\num{11543865}} & \textbf{\num{5531951}} & \num{4961943} & \num{5175757} & \textbf{\num{6626167}} & \textbf{\num{5332955}} \\
\hline
\multicolumn{5}{l}{Nvidia RTX 5090 32GB (Desktop, 575W, WebGPU), no ray-tracing hardware acceleration} & \multicolumn{5}{c}{} \\
\hline
\grayrow
  \textbf{HyperDrone} Low & \num{4096} & \textbf{\num{417403}} & \textbf{\num{409527}} & \textbf{\num{420788}} & \textbf{\num{415395}} & \textbf{\num{328219}} & \textbf{\num{321124}} & \textbf{\num{331062}} & \textbf{\num{328131}} \\
  \textbf{HyperDrone} Medium & \num{4096} & \num{410292} & \num{408345} & \textbf{\num{420788}} & \textbf{\num{415395}} & \num{295617} & \num{293244} & \textbf{\num{331062}} & \textbf{\num{328131}} \\
\grayrow
  \textbf{HyperDrone} High & \num{4096} & \num{404513} & \num{401298} & \textbf{\num{420788}} & \textbf{\num{415395}} & \num{279753} & \num{279885} & \textbf{\num{331062}} & \textbf{\num{328131}} \\
\hline
\multicolumn{5}{l}{Apple Mac Studio M5 Ultra 256GB (Desktop, 600W, Metal)} & \multicolumn{5}{c}{} \\
\hline
\grayrow
  \textbf{HyperDrone} Low & \num{4096} & \textbf{\num{5827688}} & \textbf{\num{5434048}} & \textbf{\num{5460423}} & \textbf{\num{5106184}} & \textbf{\num{2314351}} & \textbf{\num{2226623}} & \textbf{\num{2321545}} & \textbf{\num{2231347}} \\
  \textbf{HyperDrone} Medium & \num{4096} & \num{4776278} & \num{4486263} & \textbf{\num{5460423}} & \textbf{\num{5106184}} & \num{2064450} & \num{1993895} & \textbf{\num{2321545}} & \textbf{\num{2231347}} \\
\grayrow
  \textbf{HyperDrone} High & \num{4096} & \num{4527656} & \num{4275640} & \textbf{\num{5460423}} & \textbf{\num{5106184}} & \num{1500309} & \num{1451156} & \textbf{\num{2321545}} & \textbf{\num{2231347}} \\
\hline
\multicolumn{5}{l}{Nvidia H200 141GB (Server, 700W, OptiX), no ray-tracing hardware acceleration} & \multicolumn{5}{c}{} \\
\hline
\grayrow
  \textbf{HyperDrone} Low & \num{4096} & \textbf{\num{2011810}} & \textbf{\num{1994000}} & \textbf{\num{1971677}} & \textbf{\num{1955133}} & \textbf{\num{708266}} & \textbf{\num{706011}} & \textbf{\num{712375}} & \textbf{\num{709309}} \\
  \textbf{HyperDrone} Medium & \num{4096} & \num{1756810} & \num{1743380} & \textbf{\num{1971677}} & \textbf{\num{1955133}} & \num{510551} & \num{509468} & \textbf{\num{712375}} & \textbf{\num{709309}} \\
\grayrow
  \textbf{HyperDrone} High & \num{4096} & \num{1691910} & \num{1679590} & \textbf{\num{1971677}} & \textbf{\num{1955133}} & \num{373132} & \num{343078} & \textbf{\num{712375}} & \textbf{\num{709309}} \\
\hline
\end{tabular}
}
\caption{Simulator Throughput Comparison. Resolution: 64 $\times$ 64, Unit: Frames Per Second (FPS). OOM = out of memory, T/O = timeout $>15$ min.}
\label{tab:benchmark_results}
\vspace{-10pt}
\end{table*}

\subsection{Performance}

We find that HyperDrone achieves exceptional rendering and full simulation performance. Table \ref{tab:benchmark_results} shows a benchmark comparison of simulator throughput for existing simulators/renderers and HyperDrone. We benchmark in two scenarios: a simple scene with 20 primitive objects and the full indoor scene that is also seen in Figures \ref{fig:procthor}, \ref{fig:fidelity}, and \ref{fig:motion-blur}. For each scene we also benchmark combinations of RGB/Depth and rendering-only/rendering\&physics simulation (physics on CUDA for OptiX and on one CPU thread otherwise). Vulkan renders at throughput comparable to OptiX on the RTX 5090, Metal achieves \num{1500309} FPS on a Mac Studio and \num{97376} FPS on an iPhone 17 Pro Max (ProcTHOR, RGB, High), and WebGPU is $9$--$27\times$ slower than OptiX on the same GPU. 

Since, as shown in Table \ref{tab:simulator_comparison}, some simulators are not compatible with the latest Nvidia Blackwell/5000-series GPUs, we run the benchmarks across all simulators on a RTX 4090 16GB (80W). We can see that even considering the same hardware, HyperDrone achieves more than $20\times$ the throughput of the fastest quadrotor simulators. Madrona \cite{rosenzweig2024madrona} is much faster than any of the quadrotor simulators at rendering but is still more than $5\times$ slower than HyperDrone at comparable fidelity. By comparing Figure \ref{fig:fidelity} with Fig. 3 (panel ``ProcTHOR'') of \cite{rosenzweig2024madrona} we can see that Madrona falls between the Low and Medium fidelity. 

Furthermore, Aerial Gym does not handle arbitrary 3D mesh scenes for RGB rendering, and Madrona does not contain quadrotor dynamics simulation, hence these cells remain empty. Also, Flightmare requires proprietary, licensed software to modify the scene and does not support batch rendering. Hence, it is not included in this comparison. It has been measured to yield $\approx 230$ Frames Per Second (FPS) in \cite{li2025visfly}.

\subsection{Visual Simulation-to-Reality Transfer}
\label{sec:results-sim2real}

\begin{figure}
    \centering
    \includegraphics[width=0.87\textwidth]{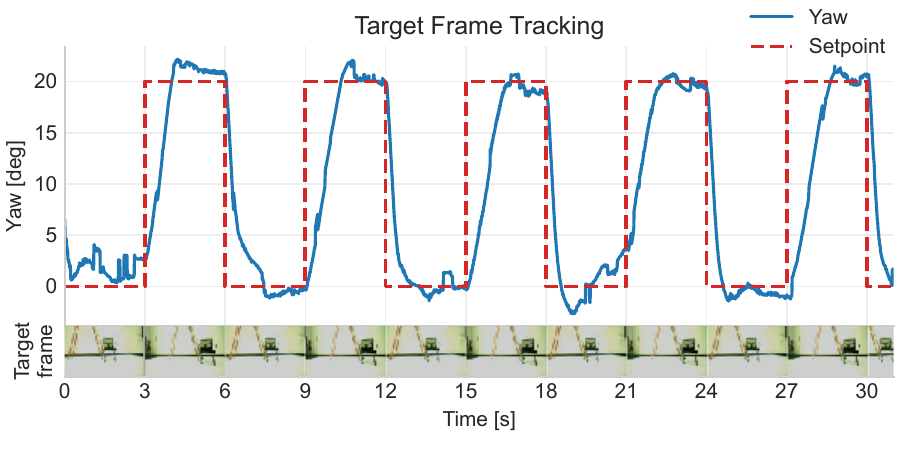}
    \vspace{-6mm}
    \caption{Real-World Target Frame Tracking: There is a chair and a ladder in the field-of-view.}
    \label{fig:yaw-tracking}
    \vspace{-4mm}
\end{figure}

The training of the visual yaw-control policy depicted takes about $4$ h on the Nvidia RTX 5090. In total the GPU process takes about $6$GB of GPU memory and hence can be run on even lower-tier consumer GPUs like an RTX 3060 8GB (in $15$ h).

We are using RLtools \cite{eschmann2024rltools} for the deep learning components because it is fast, has GPU support with fine-grained control of the memory allocations and is deterministic given a particular random seed. We also found that HyperDrone is deterministic, so multiple runs with the same seed on the same hardware and backend give exactly the same results. This is hugely beneficial for research and development as it makes it easier to reproduce bugs or other undesired or desired behaviors. 

To verify the merit of HyperDrone for visual simulation-to-reality transfer we conduct real-world experiments using the platform and setup described in Section \ref{sec:sim2real}. Figure \ref{fig:yaw-tracking} shows successful, entirely vision-based yaw tracking. Please refer to the accompanying video for an external recording of the quadrotor flying, synchronized with the onboard visual perception and the target frame.

\section{Limitations and Future Work}
While providing a step change in performance and demonstrating simulation-to-reality transfer, \mbox{HyperDrone} still has a few limitations:
\vspace{-2mm}
\begin{enumerate}
\item Rolling shutter simulation not supported, yet. The focus is on global shutter cameras but in the future we plan to support rolling shutter as well.
\item Collision detection. While probe rays are implemented, we have neither implemented a proper collision avoidance interface nor tested them for collision avoidance, yet.
\item Simple, low-frequency control problem. We only show a proof-of-concept simulation-to-reality transfer that still relies on an external position reference and RAPTOR as an end-to-end low-level controller.
\end{enumerate}
\vspace{-2mm}
Unlike related works, we built HyperDrone to be durable, maintainable and long-term stable. We plan to address the mentioned limitations and more in future research and development.

\section{Conclusion}
In this work, we presented HyperDrone, a fast and efficient simulator for synthesizing and training vision-based quadrotor control policies.
By handing this powerful simulator over to the community we expect qualitatively new results, enabling researchers and practitioners to iterate much faster and with minimal hardware requirements compared to the predominant alternatives. This lowers the barrier to do research on learned, vision-based control policies and helps to democratize this area of research. In collaboration with the community, we hope to overcome many of the limitations mentioned above and advance toward an autonomy stack that enables fully synthetic training of vision-based policies for advanced robotic autonomy. Our goal is to make such capabilities accessible on consumer-grade hardware while maintaining an open and extensible framework that can be readily explored, modified, and improved by researchers, developers, and hobbyists alike.

\label{sec:conclusion}

\clearpage
\acknowledgments{
  This work was supported by the NSF CAREER Award 2546659 and the DARPA YFA Grant D22AP00156-00.
}

\bibliography{example}

\clearpage
\appendix

\section{Multirotor Dynamics}
\label{appendix:implementation}

\begin{adjustbox}{max width=\textwidth}
\begin{minipage}{\textwidth}
    \begin{align*}
    \mathbf{s} := & \; \{\mathbf{p}, \mathbf{q}, \mathbf{v}, \boldsymbol{\omega}, \boldsymbol{\omega}_{m}\} \quad \text{(motor state: $\boldsymbol{\omega}_m \in \mathbb{R}^N$)} \\
    \dot{\mathbf{p}} = & \; \mathbf{v} \\
    \dot{\mathbf{q}} = & \; \mathbf q \circ \left[0 \;\; \boldsymbol{\omega}/ 2\right]^\top\\
    \dot{\mathbf{v}} = & \; \frac{1}{m}\left(\mathbf{\mathbf{R}}(\mathbf{q}) \left(\sum_{i=1}^N \mathbf{r}_{f_i} f_i\right) + \mathbf{f}_r\right) + \mathbf{g} \\
    f_i = & \; \sum_{j=0}^{2} K_{f_{ij}} \omega_{m_i}^j \\
    \dot{\boldsymbol{\omega}} = & \; \mathbf{J}^{-1}\left(\boldsymbol{\tau} + \boldsymbol{\tau}_r + \left(\mathbf{J} \boldsymbol{\omega}\right) \times \boldsymbol{\omega}\right) \\
    \boldsymbol{\tau} = & \; \sum_{i=1}^N\left[ \left(\mathbf{r}_{p_i} \times \mathbf{r}_{f_i}\right)f_i + \mathbf{r}_{\tau_i}K_{\tau_i} f_i \right]  \\
    \dot{\omega}_{m_i} = & \; T_{m_i}^{-1} \left(\omega_{{sp}_i} - \omega_{m_i}\right) \\
    \boldsymbol{\omega}_{sp} := & \; \mathbf{a} \\
    \end{align*}

    \begingroup
    \centering
    \def\arraystretch{1.5}
    \setlength{\tabcolsep}{1mm}
    \begin{tabular}{c l}
    \hline
    Variable & Description \\
    \hline
    \grayrow
    $N$ & Number of rotors/motors \\
    $\mathbf{p}$ / $\mathbf{v}$ & Global position / velocity \\
    \grayrow
    $\mathbf{q}$ / $\mathbf{R}(\mathbf{q})$ & Orientation: quaternion / rotation matrix \\
    ${\boldsymbol{\omega}}$ / $f_i$  & Angular rate (body frame) / thrust of motor $i$  \\
    \grayrow
    $\boldsymbol{\omega}_m$ / ${\boldsymbol{\omega}_{sp}}$ & Motor RPMs: state / setpoints \\
    ${m}$ / ${\mathbf{g}}$ & Mass / gravity \\
    \grayrow
    ${\mathbf{r}_{p_i}}$ / ${\mathbf{r}_{f_i}}$ / ${\mathbf{r}_{\tau_i}}$  & Motor $i$: position / force direction / torque direction \\
    ${\mathbf{J}}$ / ${T_{m_i}}$ & Inertia matrix / Motor $i$ delay time constant \\
    \grayrow
    ${K_{\tau_{i}}}$ / ${K_{f_{ij}}}$ & Motor $i$: torque coeff. / thrust coeff. (exponent $j$) \\
    ${\mathbf{f}_r}$ / $\boldsymbol{\tau}_r$ & Random force / random torque \\
    \hline
    \end{tabular}
    \captionof{table}{Variables and parameters of the multirotor dynamics model.}
    \endgroup

\end{minipage}
\end{adjustbox}

\section{Rendering Fidelity Modes}
\label{appendix:rendering_fidelity_modes}
HyperDrone aims to provide a controllable trade-off between fidelity and throughput. The three modes compared in Figure~\ref{fig:fidelity} share ray generation (one primary ray per anti-aliasing and motion-blur sample), the acceleration structure and the depth, segmentation, surface-normal and optical-flow outputs. The modes only differ in how an RGB hit is shaded. Each mode is a compile-time set of feature flags, so disabled features are zero-cost at runtime. Materials follow the glTF 2.0 material model, which codifies Physically Based Rendering (PBR) in ISO/IEC 12113:2022 \cite{iso12113gltf}, so any glTF asset renders without conversion. Table~\ref{tab:fidelity_modes} summarizes the modes.

\paragraph{Low.} Textures are ``baked'': Each mesh receives one flat color at load time (the material color factor multiplied by the mean color of its base-color texture), and a hit returns this color independent of surface orientation and lighting.

\paragraph{Medium.} Base-color textures are sampled at the hit point and shaded by a simple view-dependent term (surfaces darken towards grazing angles) without scene lights. Metallic materials trace one reflection ray.

\paragraph{High.} The full glTF 2.0 PBR material (base-color, metallic-roughness, normal, occlusion and emissive maps) is lit by the scene's punctual lights (directional, point and spot lights as authored in the ProcTHOR scenes) plus a constant ambient term that approximates indirect illumination ($0.6$ in the converted ProcTHOR scenes). Metallic surfaces trace one reflection ray (e.g. the reflection of the picture in the table surface in Figure~\ref{fig:fidelity}) and transparent materials one continuation ray. Not (yet) modeled are shadows, multi-bounce indirect illumination and subsurface scattering.

\begingroup
\centering
\def\arraystretch{1.3}
\setlength{\tabcolsep}{2mm}
\begin{tabular}{l c c c}
\hline
Feature & Low & Medium & High \\
\hline
\grayrow
Base-color textures & \customnocheckmark & \customcheckmark & \customcheckmark \\
Orientation-dependent shading & \customnocheckmark & geometric normal & normal map \\
\grayrow
Scene lights & \customnocheckmark & \customnocheckmark & \customcheckmark \\
Material model & flat color & color + metallic & metallic-roughness BRDF \\
\grayrow
Reflections (one bounce) & \customnocheckmark & metallic $> 0$ & metallic $> 0.1$ \\
Emissive, occlusion, transparency & \customnocheckmark & \customnocheckmark & \customcheckmark \\
\hline
\end{tabular}
\captionof{table}{Rendering Fidelity Modes.}
\label{tab:fidelity_modes}
\endgroup

\section{Training Pipeline}
\label{appendix:training_pipeline}
The yaw-estimation student is trained by supervised regression on trajectories flown by the RAPTOR \cite{eschmann_2026_raptor} teacher. Simulation, rendering, teacher and student run in one process on one GPU. Table~\ref{tab:training_pipeline} lists the settings/parameters and the following describes one epoch.

\paragraph{Scenes and target frames.} Of the first $30$ ProcTHOR training set scenes (\texttt{ProcTHOR-Train-\char`\{0-29\char`\}}, no cherry-picking), $25$ are used for training and $5$ are held out for validation. Every epoch draws two distinct training scenes and simulates $64$ instances in each, i.e. $128$ environments. For each instance a \emph{setup} is sampled: a position from the scene's pre-sampled list of free-space positions (minimum clearance $1$\,m, based on probe rays), a uniform scene yaw, a perturbed camera (field of view, mount offset and rotation), a brightness scale, and a roll/pitch perturbation. The $64$ target frames of a scene are rendered in one batched pass from the level attitude at the setup pose and cached for the epoch. Each instance adopts its setup, additionally samples the thrust-to-weight ratio and the shutter fraction, and is spawned at the setup position with zero linear velocity, a uniformly random angular velocity and an orientation rotated by $\theta \sim \mathcal{U}(0, \theta_\mathrm{max})$ about a uniformly random axis. Current and target camera therefore share intrinsics and mount and initially differ by this rotation only.

\paragraph{Rollout and labels.} The teacher is the RAPTOR checkpoint (Dense $16$ $\to$ GRU $16$ $\to$ Dense $4$, GRU state reset with the instance) acting on the privileged $22$-dimensional state (position, rotation matrix, linear and angular velocity, last action) and emitting direct motor commands at the $100$\,Hz simulation rate. It regulates the quadrotor back to the spawn position and the level, zero-yaw attitude, so the $50$ steps ($0.5$\,s) of an epoch sweep the yaw offset from its initial value towards zero under realistic attitude and motion-blur transients. An instance is terminated when it leaves a $1$\,m box around the spawn position or exceeds $10$\,m/s or $35$\,rad/s on any axis. The reset adopts the next cached setup with freshly sampled dynamics. The label of every frame is the body yaw $\psi$ relative to the target frame.

\paragraph{Student and optimization.} The student input is the $50 \times 80 \times 8$ stack of the current RGB frame, the target RGB frame and two zero channels (padding for the NPU). Four $3 \times 3$, stride-$2$ convolutions with $32$, $64$, $128$ and $256$ channels (ReLU) reduce it to $4 \times 5 \times 256 = 5120$ features, followed by three dense layers of $64$ units (ReLU) and a linear $2$-unit output, $0.73$\,M parameters in total.

\paragraph{Curriculum and validation.} The initial-orientation bound $\theta_\mathrm{max}$ follows a curriculum from $0^\circ$ to $90^\circ$ in $1^\circ$ increments. Every $10$ epochs the student is validated on the held-out scenes with co-located current and target frames at $11$ fixed yaw offsets up to $\pm 32^\circ$ (nominal camera and brightness) and on a \emph{frontier} band of offsets just beyond the current bound. The bound is increased when the frontier RMSE drops below $25\,\%$ of the bound (clipped to $[2.5^\circ, 5^\circ]$), so the student only encounters larger initial rotations once it resolves the current ones.

\begingroup
\centering
\def\arraystretch{1.5}
\setlength{\tabcolsep}{1.5mm}
\begin{tabular}{l l l}
\hline
Group & Parameter & Value \\
\hline
\grayrow
\textbf{Rendering} & Resolution / field of view & $80 \times 50$\,px / $79.6^\circ$ \\
 & Shading / anti-aliasing / motion blur & PBR / $2 \times 2$ grid / $2$ samples \\
\grayrow
 & Shutter fraction of the $10$\,ms frame & $\mathcal{U}(0.25, 1)$ \\
\textbf{Setup} (per target frame) & Field of view & $\pm 5^\circ$ \\
\grayrow
 & Camera mount offset / rotation & $\pm 1$\,cm / $\pm 5^\circ$ per axis \\
 & Brightness scale & $\mathcal{U}(0.5, 1.5)$ \\
\grayrow
 & Target roll / pitch & $\pm 10^\circ$ \\
 & Target exposure mismatch & $\mathcal{U}(0.75, 1.25)$ \\
\grayrow
\textbf{Instance} (per reset) & Thrust-to-weight ratio & $\mathcal{U}(1.7, 2.0)$ \\
 & Initial angular velocity & $\mathcal{U}(-1, 1)$\,rad/s per axis \\
\grayrow
 & Initial orientation & angle $\mathcal{U}(0, \theta_\mathrm{max})$, uniform axis \\
\textbf{Optimizer} & Adam $\eta$ / $\beta_1$ / $\beta_2$ / $\epsilon$ & $10^{-3}$ / $0.9$ / $0.999$ / $10^{-8}$ \\
\hline
\end{tabular}
\captionof{table}{Training Pipeline Parameters.}
\label{tab:training_pipeline}
\endgroup

\section{Collision Queries}
\label{appendix:collision_queries}
Collision queries reuse the ray-tracing pipeline: besides its image rays, every camera traces $N$ \emph{probe rays} (Figure~\ref{fig:probe-rays}) from its position against the same acceleration structure and reads back, per probe, the distance to the first hit and a hit flag (zero-copy, like the image outputs). Probe $0$ follows the camera's principal axis; the remaining probes $i = 1, \dots, N-1$ follow a fixed Fibonacci spiral on the unit sphere in the camera frame ($x$ forward, $y$ left, $z$ up), so the pattern turns with the camera,
\begin{equation*}
\mathbf{d}_i = \left(\sin\psi_i \cos\theta_i,\; \sin\psi_i \sin\theta_i,\; \cos\psi_i\right), \quad \cos\psi_i = 1 - \frac{2 i + 1}{N}, \quad \theta_i = \frac{2\pi i}{\varphi}, \quad \varphi = \frac{1 + \sqrt{5}}{2},
\end{equation*}
i.e. equal steps in $\cos\psi$ with golden-angle increments in azimuth, which covers the sphere evenly for any $N$ and is deterministic.

The free-space annotation used to spawn drones (Appendix~\ref{appendix:training_pipeline}) is built from these probes once per scene: candidate positions are sampled inside the scene's bounding box and accepted if more than $72\,\%$ of the $64$ probes hit geometry (indoors rather than outside the house), the mean hit distance is below $45\,\%$ of the scene extent (not in a void) and the nearest hit is farther than $1$\,m (clearance). At runtime the same probes provide a per-step proximity measurement against the full scene mesh instead of a simplified collision mesh.

\section{API}
\label{appendix:api}
\renewcommand{\ttdefault}{zi4}
This appendix is the \texttt{Overview} notebook of the \texttt{hyperdrone} Python package\footnote{\url{https://colab.research.google.com/github/rl-tools/hyperdrone/blob/paper/docs/Overview.ipynb}}, executed as shown: it roughly reproduces the shot of Figure~\ref{fig:modalities} and adds two more scenes through the environment API, each an instance of four drones with the camera mounted on one of them. The renderer and the batched dynamics also work standalone; see the package README and its other notebooks.
\raggedbottom
\begin{cellgroup}
\paragraph{Scene and drone table.}
Scenes are GLB files addressed by content hash (or local paths): the \texttt{Train200} catalog pins the 200 ProcTHOR training scenes, and \texttt{conta} fetches each once into a shared cache. Each drone pairs a rigged GLB with the registry\textquotesingle s dynamics model of the same name.

\begin{cellcode}{1}
\begin{Verbatim}[commandchars=\\\{\}]
\PYG{k+kn}{import}\PYG{+w}{ }\PYG{n+nn}{numpy}\PYG{+w}{ }\PYG{k}{as}\PYG{+w}{ }\PYG{n+nn}{np}
\PYG{k+kn}{from}\PYG{+w}{ }\PYG{n+nn}{hyperdrone}\PYG{+w}{ }\PYG{k+kn}{import} \PYG{n}{HyperDrone}\PYG{p}{,} \PYG{n}{Config}
\PYG{k+kn}{from}\PYG{+w}{ }\PYG{n+nn}{hyperdrone}\PYG{+w}{ }\PYG{k+kn}{import} \PYG{n}{render}
\PYG{k+kn}{from}\PYG{+w}{ }\PYG{n+nn}{hyperdrone}\PYG{n+nn}{.}\PYG{n+nn}{camera}\PYG{+w}{ }\PYG{k+kn}{import} \PYG{n}{CameraSpec}\PYG{p}{,} \PYG{n}{RGB}\PYG{p}{,} \PYG{n}{Depth}\PYG{p}{,} \PYG{n}{Segmentation}\PYG{p}{,} \PYG{n}{Normals}\PYG{p}{,} \PYG{n}{Flow}\PYG{p}{,} \PYG{n}{FlowStatus}\PYG{p}{,} \PYG{n}{ImageObservation}
\PYG{k+kn}{from}\PYG{+w}{ }\PYG{n+nn}{hyperdrone}\PYG{n+nn}{.}\PYG{n+nn}{examples}\PYG{n+nn}{.}\PYG{n+nn}{data}\PYG{+w}{ }\PYG{k+kn}{import} \PYG{n}{rigged\PYGZus{}model\PYGZus{}path}

\PYG{n}{catalog} \PYG{o}{=} \PYG{n}{render}\PYG{o}{.}\PYG{n}{datasets}\PYG{o}{.}\PYG{n}{procthor}\PYG{o}{.}\PYG{n}{Train200}\PYG{p}{(}\PYG{p}{)}
\PYG{n}{scenes} \PYG{o}{=} \PYG{p}{[}\PYG{n}{catalog}\PYG{p}{[}\PYG{l+s+sa}{f}\PYG{l+s+s2}{\PYGZdq{}}\PYG{l+s+s2}{ProcTHOR\PYGZhy{}Train\PYGZhy{}}\PYG{l+s+si}{\PYGZob{}}\PYG{n}{index}\PYG{l+s+si}{\PYGZcb{}}\PYG{l+s+s2}{\PYGZdq{}}\PYG{p}{]} \PYG{k}{for} \PYG{n}{index} \PYG{o+ow}{in} \PYG{p}{(}\PYG{l+m+mi}{1}\PYG{p}{,} \PYG{l+m+mi}{99}\PYG{p}{,} \PYG{l+m+mi}{72}\PYG{p}{)}\PYG{p}{]}
\PYG{n}{MODELS} \PYG{o}{=} \PYG{p}{(}\PYG{l+s+s2}{\PYGZdq{}}\PYG{l+s+s2}{crazyflie\PYGZus{}brushless}\PYG{l+s+s2}{\PYGZdq{}}\PYG{p}{,} \PYG{l+s+s2}{\PYGZdq{}}\PYG{l+s+s2}{arpl}\PYG{l+s+s2}{\PYGZdq{}}\PYG{p}{,} \PYG{l+s+s2}{\PYGZdq{}}\PYG{l+s+s2}{x500}\PYG{l+s+s2}{\PYGZdq{}}\PYG{p}{,} \PYG{l+s+s2}{\PYGZdq{}}\PYG{l+s+s2}{savagebee\PYGZus{}pusher}\PYG{l+s+s2}{\PYGZdq{}}\PYG{p}{)}
\PYG{n}{drones} \PYG{o}{=} \PYG{p}{[}\PYG{p}{\PYGZob{}}\PYG{l+s+s2}{\PYGZdq{}}\PYG{l+s+s2}{asset}\PYG{l+s+s2}{\PYGZdq{}}\PYG{p}{:} \PYG{n}{rigged\PYGZus{}model\PYGZus{}path}\PYG{p}{(}\PYG{n}{name}\PYG{p}{)}\PYG{p}{,}
           \PYG{l+s+s2}{\PYGZdq{}}\PYG{l+s+s2}{parameters}\PYG{l+s+s2}{\PYGZdq{}}\PYG{p}{:} \PYG{p}{\PYGZob{}}\PYG{l+s+s2}{\PYGZdq{}}\PYG{l+s+s2}{dynamics}\PYG{l+s+s2}{\PYGZdq{}}\PYG{p}{:} \PYG{p}{\PYGZob{}}\PYG{l+s+s2}{\PYGZdq{}}\PYG{l+s+s2}{dynamics}\PYG{l+s+s2}{\PYGZdq{}}\PYG{p}{:} \PYG{p}{\PYGZob{}}\PYG{l+s+s2}{\PYGZdq{}}\PYG{l+s+s2}{model}\PYG{l+s+s2}{\PYGZdq{}}\PYG{p}{:} \PYG{n}{name}\PYG{p}{\PYGZcb{}}\PYG{p}{\PYGZcb{}}\PYG{p}{\PYGZcb{}}\PYG{p}{\PYGZcb{}}
          \PYG{k}{for} \PYG{n}{name} \PYG{o+ow}{in} \PYG{n}{MODELS}\PYG{p}{]}
\end{Verbatim}
\end{cellcode}
\end{cellgroup}

\begin{cellgroup}
\paragraph{Environment.}
\texttt{Config} holds the compile-time settings from which the environment is JIT-compiled. Each scene gets its own world with one instance of four mutually visible agents; agent 0 carries the only camera, \texttt{front}, mounted at the pose of Figure~\ref{fig:modalities}.

\begin{cellcode}{2}
\begin{Verbatim}[commandchars=\\\{\}]
\PYG{n}{config} \PYG{o}{=} \PYG{n}{Config}\PYG{p}{(}\PYG{n}{preset}\PYG{o}{=}\PYG{l+s+s2}{\PYGZdq{}}\PYG{l+s+s2}{x500\PYGZus{}fpv}\PYG{l+s+s2}{\PYGZdq{}}\PYG{p}{,} \PYG{n}{num\PYGZus{}agents}\PYG{o}{=}\PYG{l+m+mi}{4}\PYG{p}{,} \PYG{n}{instances\PYGZus{}per\PYGZus{}world}\PYG{o}{=}\PYG{l+m+mi}{1}\PYG{p}{,}
                \PYG{n}{num\PYGZus{}worlds}\PYG{o}{=}\PYG{n+nb}{len}\PYG{p}{(}\PYG{n}{scenes}\PYG{p}{)}\PYG{p}{,} \PYG{n}{width}\PYG{o}{=}\PYG{l+m+mi}{1024}\PYG{p}{,} \PYG{n}{height}\PYG{o}{=}\PYG{l+m+mi}{1024}\PYG{p}{,}
                \PYG{n}{control}\PYG{o}{=}\PYG{l+s+s2}{\PYGZdq{}}\PYG{l+s+s2}{motor}\PYG{l+s+s2}{\PYGZdq{}}\PYG{p}{,} \PYG{n}{actuation}\PYG{o}{=}\PYG{l+s+s2}{\PYGZdq{}}\PYG{l+s+s2}{rotors}\PYG{l+s+s2}{\PYGZdq{}}\PYG{p}{,}
                \PYG{n}{motion\PYGZus{}blur\PYGZus{}samples}\PYG{o}{=}\PYG{l+m+mi}{16}\PYG{p}{,} \PYG{n}{dataset}\PYG{o}{=}\PYG{l+s+s2}{\PYGZdq{}}\PYG{l+s+s2}{procthor}\PYG{l+s+s2}{\PYGZdq{}}\PYG{p}{,}
                \PYG{n}{cameras}\PYG{o}{=}\PYG{p}{(}\PYG{n}{CameraSpec}\PYG{p}{(}\PYG{l+s+s2}{\PYGZdq{}}\PYG{l+s+s2}{front}\PYG{l+s+s2}{\PYGZdq{}}\PYG{p}{,} \PYG{n}{outputs}\PYG{o}{=}\PYG{p}{(}
                       \PYG{n}{RGB}\PYG{p}{(}\PYG{l+s+s2}{\PYGZdq{}}\PYG{l+s+s2}{float32}\PYG{l+s+s2}{\PYGZdq{}}\PYG{p}{)}\PYG{p}{,} \PYG{n}{Depth}\PYG{p}{(}\PYG{p}{)}\PYG{p}{,} \PYG{n}{Segmentation}\PYG{p}{(}\PYG{n}{labels}\PYG{o}{=}\PYG{l+s+s2}{\PYGZdq{}}\PYG{l+s+s2}{instance}\PYG{l+s+s2}{\PYGZdq{}}\PYG{p}{)}\PYG{p}{,}
                       \PYG{n}{Segmentation}\PYG{p}{(}\PYG{n}{labels}\PYG{o}{=}\PYG{l+s+s2}{\PYGZdq{}}\PYG{l+s+s2}{semantic}\PYG{l+s+s2}{\PYGZdq{}}\PYG{p}{)}\PYG{p}{,} \PYG{n}{Normals}\PYG{p}{(}\PYG{p}{)}\PYG{p}{,} \PYG{n}{Flow}\PYG{p}{(}\PYG{p}{)}\PYG{p}{,} \PYG{n}{FlowStatus}\PYG{p}{(}\PYG{p}{)}\PYG{p}{)}\PYG{p}{)}\PYG{p}{,}\PYG{p}{)}\PYG{p}{,}
                   \PYG{n}{observation}\PYG{o}{=}\PYG{n}{ImageObservation}\PYG{p}{(}\PYG{n}{cameras}\PYG{o}{=}\PYG{p}{(}\PYG{l+s+s2}{\PYGZdq{}}\PYG{l+s+s2}{front}\PYG{l+s+s2}{\PYGZdq{}}\PYG{p}{,}\PYG{p}{)}\PYG{p}{)}\PYG{p}{)}
\PYG{n}{mount} \PYG{o}{=} \PYG{p}{\PYGZob{}}\PYG{l+s+s2}{\PYGZdq{}}\PYG{l+s+s2}{position}\PYG{l+s+s2}{\PYGZdq{}}\PYG{p}{:} \PYG{p}{[}\PYG{o}{\PYGZhy{}}\PYG{l+m+mf}{0.2102}\PYG{p}{,} \PYG{l+m+mf}{0.2639}\PYG{p}{,} \PYG{l+m+mf}{0.3870}\PYG{p}{]}\PYG{p}{,}  \PYG{c+c1}{\PYGZsh{} Figure 1\PYGZsq{}s camera in}
         \PYG{l+s+s2}{\PYGZdq{}}\PYG{l+s+s2}{orientation}\PYG{l+s+s2}{\PYGZdq{}}\PYG{p}{:} \PYG{p}{[}\PYG{l+m+mf}{0.6276}\PYG{p}{,} \PYG{l+m+mf}{0.2865}\PYG{p}{,} \PYG{l+m+mf}{0.2584}\PYG{p}{,} \PYG{o}{\PYGZhy{}}\PYG{l+m+mf}{0.6763}\PYG{p}{]}\PYG{p}{\PYGZcb{}}  \PYG{c+c1}{\PYGZsh{} agent 0 body frame}
\PYG{n}{drones}\PYG{p}{[}\PYG{l+m+mi}{0}\PYG{p}{]}\PYG{p}{[}\PYG{l+s+s2}{\PYGZdq{}}\PYG{l+s+s2}{parameters}\PYG{l+s+s2}{\PYGZdq{}}\PYG{p}{]}\PYG{p}{[}\PYG{l+s+s2}{\PYGZdq{}}\PYG{l+s+s2}{cameras}\PYG{l+s+s2}{\PYGZdq{}}\PYG{p}{]} \PYG{o}{=} \PYG{p}{\PYGZob{}}\PYG{l+s+s2}{\PYGZdq{}}\PYG{l+s+s2}{front}\PYG{l+s+s2}{\PYGZdq{}}\PYG{p}{:} \PYG{p}{\PYGZob{}}\PYG{l+s+s2}{\PYGZdq{}}\PYG{l+s+s2}{fov}\PYG{l+s+s2}{\PYGZdq{}}\PYG{p}{:} \PYG{l+m+mi}{80}\PYG{p}{,} \PYG{l+s+s2}{\PYGZdq{}}\PYG{l+s+s2}{mount}\PYG{l+s+s2}{\PYGZdq{}}\PYG{p}{:} \PYG{p}{\PYGZob{}}\PYG{l+s+s2}{\PYGZdq{}}\PYG{l+s+s2}{pose}\PYG{l+s+s2}{\PYGZdq{}}\PYG{p}{:} \PYG{n}{mount}\PYG{p}{\PYGZcb{}}\PYG{p}{\PYGZcb{}}\PYG{p}{\PYGZcb{}}
\end{Verbatim}
\end{cellcode}
\end{cellgroup}

\begin{cellgroup}
\paragraph{Initialization.}
Runtime parameters are a partial dict merged over the nominal values. Here the drones start with zero linear velocity, propellers at full speed and a random angular velocity of up to 1 rad/s per axis, which the flow tiles show.

\begin{cellcode}{3}
\begin{Verbatim}[commandchars=\\\{\}]
\PYG{n}{initial} \PYG{o}{=} \PYG{p}{\PYGZob{}}\PYG{l+s+s2}{\PYGZdq{}}\PYG{l+s+s2}{guidance}\PYG{l+s+s2}{\PYGZdq{}}\PYG{p}{:} \PYG{l+m+mi}{0}\PYG{p}{,} \PYG{l+s+s2}{\PYGZdq{}}\PYG{l+s+s2}{max\PYGZus{}angular\PYGZus{}velocity}\PYG{l+s+s2}{\PYGZdq{}}\PYG{p}{:} \PYG{l+m+mi}{1}\PYG{p}{,} \PYG{l+s+s2}{\PYGZdq{}}\PYG{l+s+s2}{max\PYGZus{}linear\PYGZus{}velocity}\PYG{l+s+s2}{\PYGZdq{}}\PYG{p}{:} \PYG{l+m+mf}{0.0}\PYG{p}{,}
           \PYG{l+s+s2}{\PYGZdq{}}\PYG{l+s+s2}{min\PYGZus{}rpm}\PYG{l+s+s2}{\PYGZdq{}}\PYG{p}{:} \PYG{l+m+mi}{1}\PYG{p}{,} \PYG{l+s+s2}{\PYGZdq{}}\PYG{l+s+s2}{max\PYGZus{}rpm}\PYG{l+s+s2}{\PYGZdq{}}\PYG{p}{:} \PYG{l+m+mi}{1}\PYG{p}{\PYGZcb{}}
\PYG{n}{parameters} \PYG{o}{=} \PYG{p}{\PYGZob{}}\PYG{l+s+s2}{\PYGZdq{}}\PYG{l+s+s2}{dynamics}\PYG{l+s+s2}{\PYGZdq{}}\PYG{p}{:} \PYG{p}{\PYGZob{}}\PYG{l+s+s2}{\PYGZdq{}}\PYG{l+s+s2}{mdp}\PYG{l+s+s2}{\PYGZdq{}}\PYG{p}{:} \PYG{p}{\PYGZob{}}\PYG{l+s+s2}{\PYGZdq{}}\PYG{l+s+s2}{init}\PYG{l+s+s2}{\PYGZdq{}}\PYG{p}{:} \PYG{n}{initial}\PYG{p}{\PYGZcb{}}\PYG{p}{\PYGZcb{}}\PYG{p}{\PYGZcb{}}
\PYG{n}{env} \PYG{o}{=} \PYG{n}{HyperDrone}\PYG{p}{(}\PYG{p}{[}\PYG{n}{scene}\PYG{o}{.}\PYG{n}{reference} \PYG{k}{for} \PYG{n}{scene} \PYG{o+ow}{in} \PYG{n}{scenes}\PYG{p}{]}\PYG{p}{,} \PYG{n}{config}\PYG{o}{=}\PYG{n}{config}\PYG{p}{,}
                 \PYG{n}{drones}\PYG{o}{=}\PYG{n}{drones}\PYG{p}{,} \PYG{n}{seed}\PYG{o}{=}\PYG{l+m+mi}{0}\PYG{p}{,} \PYG{n}{parameters}\PYG{o}{=}\PYG{n}{parameters}\PYG{p}{)}
\end{Verbatim}
\end{cellcode}
\end{cellgroup}

\begin{cellgroup}
\paragraph{Reset and step.}
\texttt{reset} places the scenes and drones at pre-defined spawn points in agent 0\textquotesingle s field of view, \texttt{observe} renders lazily and \texttt{step} integrates 10 ms of dynamics. Actions are normalized motor commands in \texttt{{[}-1,\ 1{]}}; here every agent applies full throttle.

\begin{cellcode}{4}
\begin{Verbatim}[commandchars=\\\{\}]
\PYG{n}{formations} \PYG{o}{=} \PYG{p}{[}  \PYG{c+c1}{\PYGZsh{} (position, yaw in degrees) per agent}
    \PYG{p}{[}\PYG{p}{(}\PYG{p}{(}\PYG{o}{\PYGZhy{}}\PYG{l+m+mf}{1.602}\PYG{p}{,} \PYG{o}{\PYGZhy{}}\PYG{l+m+mf}{3.994}\PYG{p}{,} \PYG{l+m+mf}{1.145}\PYG{p}{)}\PYG{p}{,} \PYG{l+m+mi}{65}\PYG{p}{)}\PYG{p}{,} \PYG{p}{(}\PYG{p}{(}\PYG{o}{\PYGZhy{}}\PYG{l+m+mf}{0.459}\PYG{p}{,} \PYG{o}{\PYGZhy{}}\PYG{l+m+mf}{4.226}\PYG{p}{,} \PYG{l+m+mf}{1.120}\PYG{p}{)}\PYG{p}{,} \PYG{l+m+mi}{30}\PYG{p}{)}\PYG{p}{,}
     \PYG{p}{(}\PYG{p}{(}\PYG{o}{\PYGZhy{}}\PYG{l+m+mf}{0.365}\PYG{p}{,} \PYG{o}{\PYGZhy{}}\PYG{l+m+mf}{5.578}\PYG{p}{,} \PYG{l+m+mf}{1.073}\PYG{p}{)}\PYG{p}{,} \PYG{o}{\PYGZhy{}}\PYG{l+m+mi}{35}\PYG{p}{)}\PYG{p}{,} \PYG{p}{(}\PYG{p}{(}\PYG{o}{\PYGZhy{}}\PYG{l+m+mf}{1.806}\PYG{p}{,} \PYG{o}{\PYGZhy{}}\PYG{l+m+mf}{4.461}\PYG{p}{,} \PYG{l+m+mf}{0.982}\PYG{p}{)}\PYG{p}{,} \PYG{o}{\PYGZhy{}}\PYG{l+m+mi}{20}\PYG{p}{)}\PYG{p}{]}\PYG{p}{,}
    \PYG{p}{[}\PYG{p}{(}\PYG{p}{(}\PYG{o}{\PYGZhy{}}\PYG{l+m+mf}{13.976}\PYG{p}{,} \PYG{o}{\PYGZhy{}}\PYG{l+m+mf}{12.003}\PYG{p}{,} \PYG{l+m+mf}{1.103}\PYG{p}{)}\PYG{p}{,} \PYG{l+m+mi}{30}\PYG{p}{)}\PYG{p}{,} \PYG{p}{(}\PYG{p}{(}\PYG{o}{\PYGZhy{}}\PYG{l+m+mf}{13.768}\PYG{p}{,} \PYG{o}{\PYGZhy{}}\PYG{l+m+mf}{13.273}\PYG{p}{,} \PYG{l+m+mf}{0.667}\PYG{p}{)}\PYG{p}{,} \PYG{l+m+mi}{150}\PYG{p}{)}\PYG{p}{,}
     \PYG{p}{(}\PYG{p}{(}\PYG{o}{\PYGZhy{}}\PYG{l+m+mf}{14.423}\PYG{p}{,} \PYG{o}{\PYGZhy{}}\PYG{l+m+mf}{12.652}\PYG{p}{,} \PYG{l+m+mf}{0.718}\PYG{p}{)}\PYG{p}{,} \PYG{o}{\PYGZhy{}}\PYG{l+m+mi}{100}\PYG{p}{)}\PYG{p}{,} \PYG{p}{(}\PYG{p}{(}\PYG{o}{\PYGZhy{}}\PYG{l+m+mf}{13.466}\PYG{p}{,} \PYG{o}{\PYGZhy{}}\PYG{l+m+mf}{12.420}\PYG{p}{,} \PYG{l+m+mf}{0.642}\PYG{p}{)}\PYG{p}{,} \PYG{l+m+mi}{20}\PYG{p}{)}\PYG{p}{]}\PYG{p}{,}
    \PYG{p}{[}\PYG{p}{(}\PYG{p}{(}\PYG{o}{\PYGZhy{}}\PYG{l+m+mf}{2.082}\PYG{p}{,} \PYG{o}{\PYGZhy{}}\PYG{l+m+mf}{13.649}\PYG{p}{,} \PYG{l+m+mf}{0.943}\PYG{p}{)}\PYG{p}{,} \PYG{o}{\PYGZhy{}}\PYG{l+m+mi}{139}\PYG{p}{)}\PYG{p}{,} \PYG{p}{(}\PYG{p}{(}\PYG{o}{\PYGZhy{}}\PYG{l+m+mf}{2.014}\PYG{p}{,} \PYG{o}{\PYGZhy{}}\PYG{l+m+mf}{11.854}\PYG{p}{,} \PYG{l+m+mf}{0.634}\PYG{p}{)}\PYG{p}{,} \PYG{l+m+mi}{45}\PYG{p}{)}\PYG{p}{,}
     \PYG{p}{(}\PYG{p}{(}\PYG{o}{\PYGZhy{}}\PYG{l+m+mf}{2.374}\PYG{p}{,} \PYG{o}{\PYGZhy{}}\PYG{l+m+mf}{12.688}\PYG{p}{,} \PYG{l+m+mf}{0.605}\PYG{p}{)}\PYG{p}{,} \PYG{o}{\PYGZhy{}}\PYG{l+m+mi}{150}\PYG{p}{)}\PYG{p}{,} \PYG{p}{(}\PYG{p}{(}\PYG{o}{\PYGZhy{}}\PYG{l+m+mf}{3.100}\PYG{p}{,} \PYG{o}{\PYGZhy{}}\PYG{l+m+mf}{13.230}\PYG{p}{,} \PYG{l+m+mf}{0.617}\PYG{p}{)}\PYG{p}{,} \PYG{l+m+mi}{80}\PYG{p}{)}\PYG{p}{]}\PYG{p}{]}
\PYG{n}{shots} \PYG{o}{=} \PYG{p}{[}\PYG{p}{\PYGZob{}}\PYG{l+s+s2}{\PYGZdq{}}\PYG{l+s+s2}{scene}\PYG{l+s+s2}{\PYGZdq{}}\PYG{p}{:} \PYG{n}{scene}\PYG{o}{.}\PYG{n}{reference}\PYG{p}{,} \PYG{l+s+s2}{\PYGZdq{}}\PYG{l+s+s2}{drone}\PYG{l+s+s2}{\PYGZdq{}}\PYG{p}{:} \PYG{p}{[}\PYG{l+m+mi}{0}\PYG{p}{,} \PYG{l+m+mi}{1}\PYG{p}{,} \PYG{l+m+mi}{2}\PYG{p}{,} \PYG{l+m+mi}{3}\PYG{p}{]}\PYG{p}{,}
          \PYG{l+s+s2}{\PYGZdq{}}\PYG{l+s+s2}{poses}\PYG{l+s+s2}{\PYGZdq{}}\PYG{p}{:} \PYG{p}{[}\PYG{n}{pose}\PYG{p}{(}\PYG{n}{position}\PYG{p}{,} \PYG{n}{yaw}\PYG{p}{)} \PYG{k}{for} \PYG{n}{position}\PYG{p}{,} \PYG{n}{yaw} \PYG{o+ow}{in} \PYG{n}{formation}\PYG{p}{]}\PYG{p}{,}
          \PYG{l+s+s2}{\PYGZdq{}}\PYG{l+s+s2}{scene\PYGZus{}translation}\PYG{l+s+s2}{\PYGZdq{}}\PYG{p}{:} \PYG{p}{[}\PYG{l+m+mi}{0}\PYG{p}{,} \PYG{l+m+mi}{0}\PYG{p}{,} \PYG{l+m+mi}{0}\PYG{p}{]}\PYG{p}{,} \PYG{l+s+s2}{\PYGZdq{}}\PYG{l+s+s2}{scene\PYGZus{}yaw}\PYG{l+s+s2}{\PYGZdq{}}\PYG{p}{:} \PYG{p}{[}\PYG{l+m+mi}{1}\PYG{p}{,} \PYG{l+m+mi}{0}\PYG{p}{]}\PYG{p}{,}
          \PYG{l+s+s2}{\PYGZdq{}}\PYG{l+s+s2}{shutter\PYGZus{}fraction}\PYG{l+s+s2}{\PYGZdq{}}\PYG{p}{:} \PYG{l+m+mf}{0.2}\PYG{p}{\PYGZcb{}}
         \PYG{k}{for} \PYG{n}{scene}\PYG{p}{,} \PYG{n}{formation} \PYG{o+ow}{in} \PYG{n+nb}{zip}\PYG{p}{(}\PYG{n}{scenes}\PYG{p}{,} \PYG{n}{formations}\PYG{p}{)}\PYG{p}{]}
\PYG{n}{actions} \PYG{o}{=} \PYG{n}{np}\PYG{o}{.}\PYG{n}{ones}\PYG{p}{(}\PYG{p}{(}\PYG{n}{env}\PYG{o}{.}\PYG{n}{total\PYGZus{}instances}\PYG{p}{,} \PYG{n}{env}\PYG{o}{.}\PYG{n}{action\PYGZus{}dim}\PYG{p}{)}\PYG{p}{,} \PYG{n}{dtype}\PYG{o}{=}\PYG{n}{np}\PYG{o}{.}\PYG{n}{float32}\PYG{p}{)}
\PYG{n}{env}\PYG{o}{.}\PYG{n}{reset}\PYG{p}{(}\PYG{n}{spawn\PYGZus{}points}\PYG{o}{=}\PYG{n}{shots}\PYG{p}{)}
\PYG{n}{env}\PYG{o}{.}\PYG{n}{observe}\PYG{p}{(}\PYG{p}{)}                               \PYG{c+c1}{\PYGZsh{} first capture: the flow reference}
\PYG{n}{env}\PYG{o}{.}\PYG{n}{step}\PYG{p}{(}\PYG{n}{actions}\PYG{p}{)}
\end{Verbatim}
\end{cellcode}
\end{cellgroup}

\begin{cellgroup}
\paragraph{Outputs.}
\texttt{render()} acquires the capture after the step. \texttt{env.outputs{[}descriptor{]}.read()} returns an \texttt{(instances,\ selected\_views,\ H,\ W{[},\ C{]})} snapshot and accepts \texttt{out=} buffers (CUDA arrays on OptiX); each scene is one row of the mosaic.

\begin{cellcode}{5}
\begin{Verbatim}[commandchars=\\\{\}]
\PYG{n}{env}\PYG{o}{.}\PYG{n}{render}\PYG{p}{(}\PYG{p}{)}
\PYG{n}{mosaic}\PYG{p}{(}\PYG{p}{[}\PYG{n}{scene}\PYG{o}{.}\PYG{n}{id} \PYG{k}{for} \PYG{n}{scene} \PYG{o+ow}{in} \PYG{n}{scenes}\PYG{p}{]}\PYG{p}{,} \PYG{p}{\PYGZob{}}
    \PYG{l+s+s2}{\PYGZdq{}}\PYG{l+s+s2}{RGB}\PYG{l+s+s2}{\PYGZdq{}}\PYG{p}{:} \PYG{n}{env}\PYG{o}{.}\PYG{n}{outputs}\PYG{p}{[}\PYG{n}{RGB}\PYG{p}{(}\PYG{l+s+s2}{\PYGZdq{}}\PYG{l+s+s2}{float32}\PYG{l+s+s2}{\PYGZdq{}}\PYG{p}{)}\PYG{p}{]}\PYG{o}{.}\PYG{n}{read}\PYG{p}{(}\PYG{p}{)}\PYG{p}{[}\PYG{p}{:}\PYG{p}{,} \PYG{l+m+mi}{0}\PYG{p}{]}\PYG{p}{,}
    \PYG{l+s+s2}{\PYGZdq{}}\PYG{l+s+s2}{Depth}\PYG{l+s+s2}{\PYGZdq{}}\PYG{p}{:} \PYG{n}{depth\PYGZus{}image}\PYG{p}{(}\PYG{n}{env}\PYG{o}{.}\PYG{n}{outputs}\PYG{p}{[}\PYG{n}{Depth}\PYG{p}{(}\PYG{p}{)}\PYG{p}{]}\PYG{o}{.}\PYG{n}{read}\PYG{p}{(}\PYG{p}{)}\PYG{p}{[}\PYG{p}{:}\PYG{p}{,} \PYG{l+m+mi}{0}\PYG{p}{]}\PYG{p}{)}\PYG{p}{,}
    \PYG{l+s+s2}{\PYGZdq{}}\PYG{l+s+s2}{Surface normals}\PYG{l+s+s2}{\PYGZdq{}}\PYG{p}{:} \PYG{n}{normals\PYGZus{}image}\PYG{p}{(}\PYG{n}{env}\PYG{o}{.}\PYG{n}{outputs}\PYG{p}{[}\PYG{n}{Normals}\PYG{p}{(}\PYG{p}{)}\PYG{p}{]}\PYG{o}{.}\PYG{n}{read}\PYG{p}{(}\PYG{p}{)}\PYG{p}{[}\PYG{p}{:}\PYG{p}{,} \PYG{l+m+mi}{0}\PYG{p}{]}\PYG{p}{)}\PYG{p}{,}
    \PYG{l+s+s2}{\PYGZdq{}}\PYG{l+s+s2}{Semantic segmentation}\PYG{l+s+s2}{\PYGZdq{}}\PYG{p}{:} \PYG{n}{segmentation\PYGZus{}image}\PYG{p}{(}
        \PYG{n}{env}\PYG{o}{.}\PYG{n}{outputs}\PYG{p}{[}\PYG{n}{Segmentation}\PYG{p}{(}\PYG{n}{labels}\PYG{o}{=}\PYG{l+s+s2}{\PYGZdq{}}\PYG{l+s+s2}{semantic}\PYG{l+s+s2}{\PYGZdq{}}\PYG{p}{)}\PYG{p}{]}\PYG{o}{.}\PYG{n}{read}\PYG{p}{(}\PYG{p}{)}\PYG{p}{[}\PYG{p}{:}\PYG{p}{,} \PYG{l+m+mi}{0}\PYG{p}{]}\PYG{p}{,} \PYG{n}{semantic}\PYG{o}{=}\PYG{k+kc}{True}\PYG{p}{)}\PYG{p}{,}
    \PYG{l+s+s2}{\PYGZdq{}}\PYG{l+s+s2}{Instance segmentation}\PYG{l+s+s2}{\PYGZdq{}}\PYG{p}{:} \PYG{n}{segmentation\PYGZus{}image}\PYG{p}{(}
        \PYG{n}{env}\PYG{o}{.}\PYG{n}{outputs}\PYG{p}{[}\PYG{n}{Segmentation}\PYG{p}{(}\PYG{n}{labels}\PYG{o}{=}\PYG{l+s+s2}{\PYGZdq{}}\PYG{l+s+s2}{instance}\PYG{l+s+s2}{\PYGZdq{}}\PYG{p}{)}\PYG{p}{]}\PYG{o}{.}\PYG{n}{read}\PYG{p}{(}\PYG{p}{)}\PYG{p}{[}\PYG{p}{:}\PYG{p}{,} \PYG{l+m+mi}{0}\PYG{p}{]}\PYG{p}{)}\PYG{p}{,}
    \PYG{l+s+s2}{\PYGZdq{}}\PYG{l+s+s2}{Optical flow}\PYG{l+s+s2}{\PYGZdq{}}\PYG{p}{:} \PYG{n}{flow\PYGZus{}image}\PYG{p}{(}
        \PYG{n}{env}\PYG{o}{.}\PYG{n}{outputs}\PYG{p}{[}\PYG{n}{Flow}\PYG{p}{(}\PYG{p}{)}\PYG{p}{]}\PYG{o}{.}\PYG{n}{read}\PYG{p}{(}\PYG{p}{)}\PYG{p}{[}\PYG{p}{:}\PYG{p}{,} \PYG{l+m+mi}{0}\PYG{p}{]}\PYG{p}{,} \PYG{n}{env}\PYG{o}{.}\PYG{n}{outputs}\PYG{p}{[}\PYG{n}{FlowStatus}\PYG{p}{(}\PYG{p}{)}\PYG{p}{]}\PYG{o}{.}\PYG{n}{read}\PYG{p}{(}\PYG{p}{)}\PYG{p}{[}\PYG{p}{:}\PYG{p}{,} \PYG{l+m+mi}{0}\PYG{p}{]}\PYG{p}{)}\PYG{p}{,}
\PYG{p}{\PYGZcb{}}\PYG{p}{)}
\end{Verbatim}
\end{cellcode}

\begin{cellout}
\includegraphics[width=\linewidth]{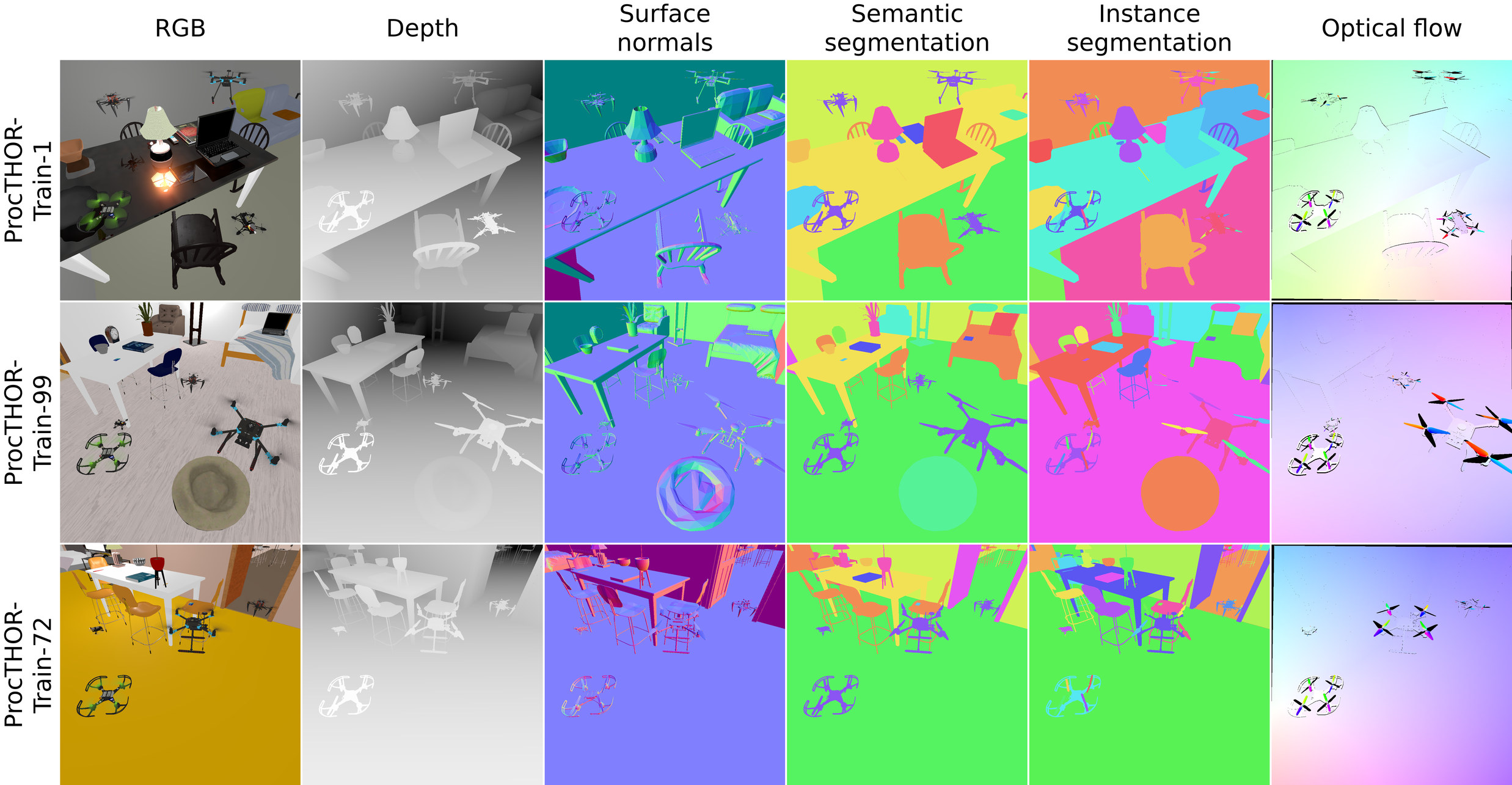}
\end{cellout}
\end{cellgroup}

\end{document}